\documentclass{article} % For LaTeX2e
\UseRawInputEncoding

\usepackage{iclr2023_conference,times}

\usepackage{amsmath,amsfonts,bm}

\def\eqref#1{equation~\ref{#1}}
\def\1{\bm{1}}

\DeclareMathAlphabet{\mathsfit}{\encodingdefault}{\sfdefault}{m}{sl}
\SetMathAlphabet{\mathsfit}{bold}{\encodingdefault}{\sfdefault}{bx}{n}

\usepackage{hyperref}
\usepackage{url}
\usepackage{graphicx}
\usepackage{multirow}
\usepackage{booktabs}
\usepackage{listings}
\usepackage{algorithm}
\usepackage{algpseudocode}
\usepackage{subfigure}

\title{When Giant Language Brains Just Aren't Enough! Domain Pizzazz with Knowledge Sparkle Dust}

\author{Minh-Tien Nguyen$^{1,2}$, Duy-Hung Nguyen$^1$, Shahab Sabahi$^1$, Hung Le$^3$, Jeff Yang$^1$, Hajime Hotta$^1$ \\
$^1$ Cinnamon AI, 10th floor, Geleximco building, 
36 Hoang Cau, Dong Da, Hanoi, Vietnam.\\
\texttt{\{ryan.nguyen, hector, sshahab, jeff.yang, hajime\}@cinnamon.is} \\
$^2$ Hung Yen University of Technology and Education, Vietnam.\\
\texttt{tiennm@utehy.edu.vn}\\
$^3$ Deakin University, Australia.\\
\texttt{thai.le@deakin.edu.au}
}

\iclrfinalcopy % Uncomment for camera-ready version, but NOT for submission.
\begin{document}

\maketitle

\begin{abstract}
%Large language models (LLMs) have achieved promising results in many natural language process tasks. However, the adaptation of LLMs to business scenarios is still an open question. This paper provides an empirical analysis to fill the gap of how to adapt LLMs to actual cases. 
Large language models (LLMs) have significantly advanced the field of natural language processing, with GPT models at the forefront. While their remarkable performance spans a range of tasks, adapting LLMs for real-world business scenarios still poses challenges warranting further investigation. This paper presents an empirical analysis aimed at bridging the gap in adapting LLMs to practical use cases. 
To do that, we select the question answering (QA) task of insurance as a case study due to its challenge of reasoning. Based on the task we design a new model relied on LLMs which are empowered by additional knowledge extracted from insurance policy rulebooks and DBpedia. The additional knowledge helps LLMs to understand new concepts of insurance for domain adaptation. Preliminary results on two QA datasets show that knowledge enhancement significantly improves the reasoning ability of GPT-3.5 (55.80\% and 57.83\% in terms of accuracy). The analysis also indicates that existing public knowledge bases, e.g., DBPedia is beneficial for knowledge enhancement.
Our findings reveal that the inherent complexity of business scenarios often necessitates the incorporation of domain-specific knowledge and external resources for effective problem-solving.

\end{abstract}

\section{Introduction}

% The introduction of LLMs, espectiall GPT and ChatGPT

% How LLMs change NLP research and market.
As the field of artificial intelligence (AI) continues to advance at an unprecedented pace, Large Language Models (LLMs)\footnote{We use the concept of LLMs as large language models with billion parameters compared to pre-trained models such as BERT \citep{DCLT-NAACL-19} with million parameters.} \citep{brown2020language,lieber2021jurassic,rae2021scaling,thoppilan2022lamda,smith2022using} and Generative Pre-trained Transformers (GPT) \citep{radford2018improving,brown2020language} have emerged as breakthrough technologies, revolutionizing natural language processing (NLP) tasks. These powerful models have demonstrated remarkable capabilities in various use cases, ranging from AI assistants, chatbots, programming, gaming, question answering, recommendation systems, and retrieval, to knowledge discovery \citep{9835523}. However, while LLMs (we use the same concept of LLMs and GPT as LLMs) offer immense potential, they are not standalone solutions \citep{iNT}, and their full effectiveness can be realized through synergies with other AI-powered technologies \citep{9835523} and tools \citep{lu2023bounding}.

In this paper, we will delve into the compelling use cases of LLMs \citep{radford2018improving,brown2020language}, showcasing their capabilities and impact in different domains. We will highlight their strengths and limitations, and examine how they can be complemented by other cutting-edge AI technologies to achieve comprehensive and powerful solutions. Through detailed firsthand findings from our internal research projects, we will shed light on the synergistic potential of integrating LLMs with complementary AI-powered tools and techniques.

We aim to provide a comprehensive overview of the use cases of LLMs, and highlight how these advanced language models can be harnessed in conjunction with other AI technologies to unlock their full potential. By leveraging the collective power of multiple AI tools and techniques, we can pave the way for innovations and solutions that can revolutionize industries and drive unprecedented advancements in the field of artificial intelligence.

Embark with us on this rigorous intellectual exploration, as we delve into the synergistic potential of LLMs and investigate the integration of these state-of-the-art technologies with complementary AI-driven tools to achieve comprehensive and highly effective solutions. Our aim is to unlock the full potential of artificial intelligence by harnessing the combined strengths of LLMs, GPT, and other sophisticated AI methodologies within a cohesive framework. 

This paper offers an analysis of the current state of research on LLMs and their applications, drawing upon a range of relevant sources and methodologies.

% Historical overview of the technical advancements that have contributed to LLMs :  progression
\subsection{Technical Advancements}
There are a lot of technical advancements relating to LLMs. This section historically reviews concepts and techniques that are directly relevant to LLMs.

% 1. Language models
\paragraph{Language modeling}
is a fundamental NLP task that was originally designed for the probability estimation of an input sequence \citep{brown1992class,pauls2011faster,jing2019survey}. The assumption of LM is based on the probability of the co-occurrence of tokens observed on a huge corpus. Given an input sequence $S = \{w_1, w_2, ..., w_n\}$ with $n$ tokens, the probability of $S$ is computed as a sequence of conditional probabilities as $P(S) = \prod_i^n P(w_i|w_1, ..., w_{i-1})$.
% \begin{equation}
%     P(S) = P(w_1, w_2, ..., w_n) = P(w_1)P(w_2|w_1)P(w_3|w_1, w_2)...P(w_n|w_1, ..., w_{n-1})
% \end{equation}
% \begin{equation}\label{eq:p-lm}
%     P(S) = P(w_1, w_2, ..., w_n) = \prod_i^n P(w_i|w_1, ..., w_{i-1})
% \end{equation}
The computation shows the nature of language modeling that can predict the current token $w_n$ given $n-1$ previous tokens. However, the computation faces the problem that the probability of a long sequence is asymptotically zero, so the probability of the whole sequence $S$ tends to be zero. To avoid this problem, language modeling uses the naive Markov assumption \citep{lauritzen1990independence}. Instead of conditionally depending on the whole $n-1$ previous tokens, the naive Markov assumption states that the probability appearance of the current token $w_i$ only conditions by $k$ previous tokens.
% \begin{equation}
%     P(S) = P(w_1, w_2, ..., w_n) \approx \prod_i P(w_i|w_{i-k}, ..., w_{i-1})
% \end{equation}

\begin{figure}[!h]
    \centering
    \includegraphics[width=0.75\textwidth]{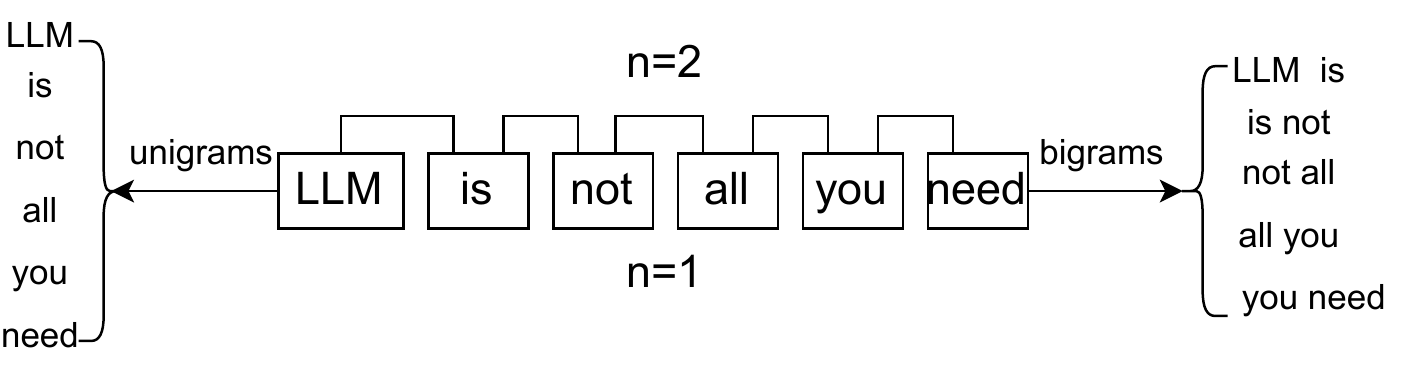}
    \caption{The example of a bi-grams LM.}
    \label{fig:n-gram}
\end{figure}

This leverages the definition of $n$-grams language models (LMs) \citep{brown1992class}. For example, the uni-gram LM only considers the probability of the current token $w_i$.
% \begin{equation}
%     P(S) = P(w_1, w_2, ..., w_n) \approx \prod_i P(w_i)
% \end{equation}
The bi-grams LM estimates the probability of the current token $w_i$ given the previous token $w_{i-1}$.
% \begin{equation}
%     P(S) = P(w_1, w_2, ..., w_n) \approx \prod_i P(w_i|w_{i-1})
% \end{equation}
The tri-grams LM estimates the probability of the current token $w_i$ given two previous tokens $w_{i-1}$ and $w_{i-2}$.
% \begin{equation}
%     P(S) = P(w_1, w_2, ..., w_n) \approx \prod_i P(w_i|w_{i-1}, w_{i-2})
% \end{equation}
Figure \ref{fig:n-gram} shows an example of a bi-grams language model. Given an input sequence \textit{``LLM is not all you need"}, the uni-gram LM ($n=1$) estimates the probability of single tokens and the bi-gram LM ($n=2$) computes the probability of two consecutive tokens. $n$-grams LMs play an important role for many NLP problems such as machine translation, error checking and correction, or text generation. For example, $n$-grams LMs can be used to give a higher probability of a sequence \textit{``LLM is not all you need"} than a sequence \textit{``LLM are not all you need"} due to the inconsistency of the subject and the verb of the second sequence. This is because the bi-gram LM returns a higher probability of the bi-gram \textit{``LLM is"} than the bi-gram \textit{``LLM are"} based on the statistics over a huge training corpus.

The concept of $n$-grams LMs leverages a lot of studies to build LMs by using neural networks \citep{bengio2000neural,mikolov2010recurrent,sundermeyer2012lstm}. Recently, the paradigm shift has also shown that all NLP tasks can be formulated as text generation handled by LLMs \citep{radford2018improving,brown2020language,Gururangan-Dont-stop-pre-training-ACL-20,Du-All-NLP-arXiv-21}.\footnote{The evaluation and smoothing methods of $n$-grams LMs are out-of-scope of this report.} However, the adaptation of LLMs to practical applications is a non-trivial task due to the challenges stated in Section \ref{sec:limitations}.

%For training, probabilities of $n$-grams LMs\footnote{https://github.com/kpu/kenlm} are estimated by using a large corpus by using likelihood estimation. 

% 2. Seq2seq and attention
\paragraph{Seq2seq and attention}
The sequence-to-sequence (seq2seq) architecture was originally designed for neural machine translation (NMT) \citep{sutskever2014sequence}. The architecture consists of two components: an encoder and a decoder. The encoder encodes an input sequence into a fixed-length context vector $\boldsymbol{V} \in \mathbb{R}^k$ and the decoder uses the vector $\boldsymbol{V}$ for decoding output tokens.
\begin{figure}[!h]
    \centering
    \includegraphics[width=0.85\textwidth]{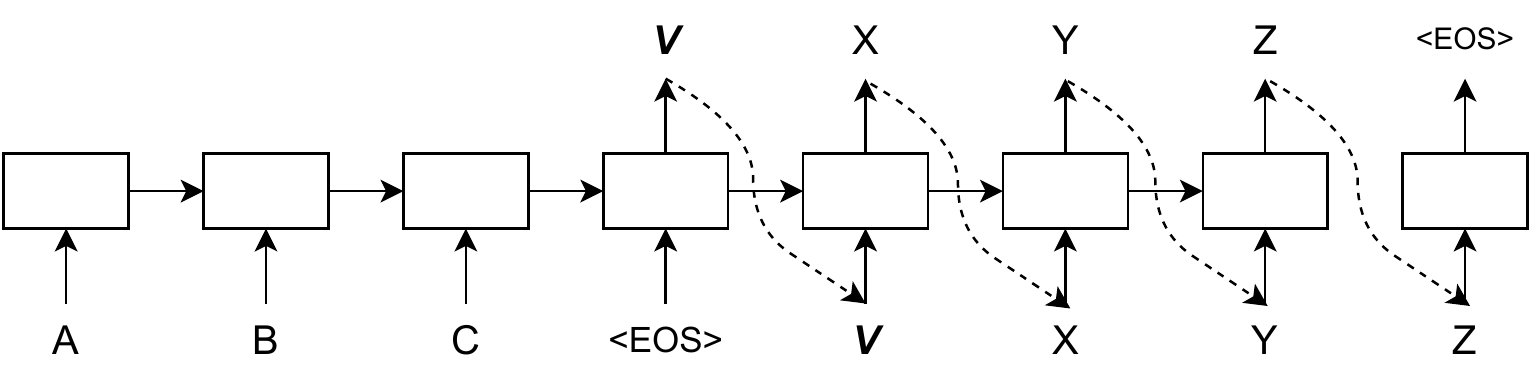}
    \caption{The architecture of seq2seq.}
    \label{fig:seq2seq}
\end{figure}

In Figure \ref{fig:seq2seq} the encoder receives an input sequence and encodes the sequence to create the vector $\boldsymbol{V}$. The vector $\boldsymbol{V}$ is the input for the decoder that outputs the next tokens based on previous predicted tokens. The decoding stops when the decoder outputs the end-of-sequence token. Common recurrent neural networks (RNNs) \cite{graves2013generating} and their variants (LSTM \citep{hochreiter1997long} and GRUs \citep{cho2014learning}) can be used for the encoder and decoder.

\begin{figure}[!h]
    \centering
    \includegraphics[width=0.7\textwidth]{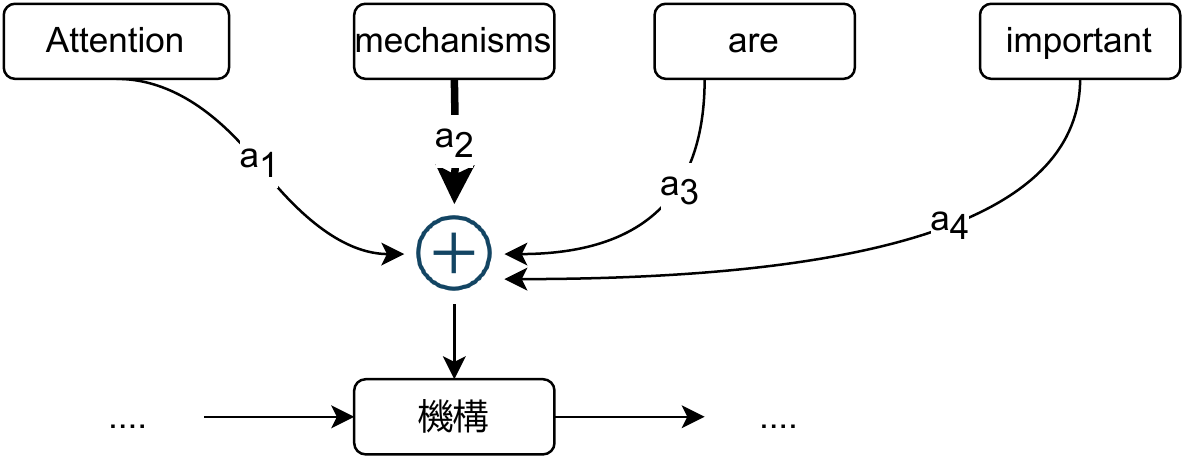}
    \caption{The attention mechanism for NMT from English to Japanese.}
    \label{fig:attn}
\end{figure}

The seq2seq architecture can be applied to several NLP problems such as NMT \citep{sutskever2014sequence}, seq2seq learning \citep{sutskever2014sequence}, text summarization \citep{chopra2016abstractive}, grammatical error correction \citep{Raffel2020ExploringTL}, incomplete utterance restoration \citep{inoue2022enhance}, etc. However, this architecture suffers in encoding long sequences that cause long-term dependency and the bottleneck of information when the whole input sequence is represented by only using the context vector in the final state of the encoder. It motivated the introduction of attention \citep{bahdanau2015neural} or pointer networks \citep{vinyals2015pointer} (we review attention because it is used in Transformer).  The core concept of attention is that during each step of the decoding process, the decoder selectively concentrates on a specific segment of the input sequence \citep{bahdanau2015neural}. This selective focus is represented by a weight matrix, which encapsulates the attention distribution across input tokens. The weight matrix is then integrated with the decoder's hidden state to facilitate the prediction of an output token for each time step, ensuring that the generated output is contextually relevant and accurately captures relationships between input tokens. Figure \ref{fig:attn} shows an example of attention of NMT.
Given an input sequence \textit{``Attention mechanisms are important"}, the attention assigns a weight $a_i$ associated with each input token. If the weight $a_i$ is high, it means that the input token $w_i$ is more important than others to predict the corresponding token in the output sequence. In essence, the attention mechanism allows the model to dynamically allocate its processing focus to the most salient parts of the input sequence at each decoding step. By adaptively directing its focus, the decoder can more effectively learn and represent the underlying structure and dependencies within the input sequence, ultimately leading to improved performance and more accurate output generation. This sophisticated approach to sequence processing ensures that the model is flexible and efficient, even when dealing with complex or lengthy input sequences.

% 3. Transformers and the lanscape pf PLMs and LLMs
\paragraph{Transformer}
The seq2seq and attention provide an effective architecture for text-to-text problems. However, the seq2seq uses recurrent models that typically factor computation in terms of memory constraint limitation and computational efficiency, especially for long sequences. It comes from the nature of sequential encoding and decoding. In addition, the attention mechanism has proven to be a highly effective technique for encoding long input sequences, thereby alleviating the information bottleneck issues that plagued earlier models \citep{parikh2016decomposable}. However, the attention mechanism is used as an additional component with a recurrent network. These two shortcomings of recurrent models and attention leverage the introduction of Transformer \citep{Vaswani-attention-NISP-17}.

The Transformer relies on the encoder-decoder architecture \citep{sutskever2014sequence}, which includes self-attention, point-wise, and fully connected layers \citep{Vaswani-attention-NISP-17}. Its main idea is to transform input data into encoded vectors with the use of self-attention. The transformation of the transformer computes attention by using query ($Q$), key ($K$), and value ($V$) matrices. We recommend audiences for reading the original paper \citep{Vaswani-attention-NISP-17} for more detailed information.
% \begin{equation}
%     Attention(Q, K, V ) = softmax(\frac{QK^T}{\sqrt{d_k}})V
% \end{equation}
% where $d_k$ is the dimension of keys, and $d_v$ is the dimension of values.
% Moreover, the Transformer uses ``multi-head attention" in parallel as the following:
% \begin{equation}
%     MultiHead(Q, K, V ) = Concat(head_1, ..., head_h)W^O
% \end{equation}
% where $head_i = Attention(QW_i^Q, KW_i^K, VW_i^V)$ in which $W_i^Q \in  \mathbb{R}^{d_{model} \times d_k}$, $W_i^K \in  \mathbb{R}^{d_{model} \times d_k}$, $W_i^V \in  \mathbb{R}^{d_{model} \times d_v}$, and $W^O \in  \mathbb{R}^{{hd_v} \times d_{model}}$. 

The success of Transformer \citep{Vaswani-attention-NISP-17} leverages the growth of pre-trained language models (PLMs) and LLMs. There are three ways for training PLMs and LLMs \citep{Amatriain-Transformer-23}.
\begin{itemize}
    \item \textbf{Encoder pretraining}: This way only uses the encoder of the Transformer for pretraining. It is also called bi-directional or auto-encoding. Models in this way are usually pre-trained by using the masked language model that reconstructs masked tokens of input sequences. In training, attention can access all input tokens. There are many pre-trained models of this way such as BERT \citep{DCLT-NAACL-19} or RoBERTa \citep{liu2019roberta}.
    
    \item \textbf{Decoder pretraining}: Different from the encoder pretraining way, the decoder of pretraining only uses the decoder of the Transformer for training PLMs and LLMs. Models are trained by predicting the next tokens (single or a set of tokens called $n$-grams). Due to their nature, these models (e.g., GPT \citep{radford2018improving}) are appropriate for text generation with promising results of many NLP tasks \citep{brown2020language}.
    
    \item \textbf{Encoder-decoder pretraining}: The third way follows the encoder-decoder of Transformer (also called as seq2seq) for pretraining. The attention of the encoder can access all input tokens while that of the decoder can only access the position of tokens before a given token. There are several PLMs of this way such as T5 \citep{Raffel2020ExploringTL}, UniLM \citep{bao2020unilmv2}, or ERNIE-GEN \citep{xiao2020ernie}.
\end{itemize}

% 4. Zero-shot learning, few-shot learning, and transfer learning
\paragraph{Transfer learning, zero-shot learning, and few-shot learning}
Transfer learning (TL) is a machine learning (ML) technique that allows ML models for transfering knowledge \citep{pan2010survey}. Different from traditional ML methods that work well under the assumption: training and testing data are drawn from the same feature space and distribution, TL provides a way that ML models can be adapted for new domains by using its existing knowledge. It usually uses fine-tuning that adapts pretrained ML models with target training samples \citep{Gururangan-Dont-stop-pre-training-ACL-20}. In business scenarios, TL can help to reduce the number of training samples \citep{nguyen2020transfer,nguyen2021transformers,nguyen2023gain}.

Zero-shot and few-shot learning are special types of TL. Zero-shot learning is a type of ML that can recognize objects whose instances may not have been seen during training \citep{larochelle2008zero,xian2017zero}. While zero-shot learning does not require out-domain samples for prediction, few-shot learning requires a few out-domain samples for training \citep{wang2020generalizing}. Based on the nature of generative models, LLMs can work well with both zero-shot and few-shot learning \citep{radford2018improving,Schick-Cloze-Question-Few,brown2020language,Liu-In-Context-GPT3-21}.

%\footnote{GPT and ChatPGT can work well with zero-shot and few-shot learning by using prompt engineering.}

% 5. HITL-RL (Toni)
%\paragraph{Human-in-the-loop and reinforcement learning}

\begin{figure}[t]
    \centering
    \includegraphics[width=0.65\textwidth]{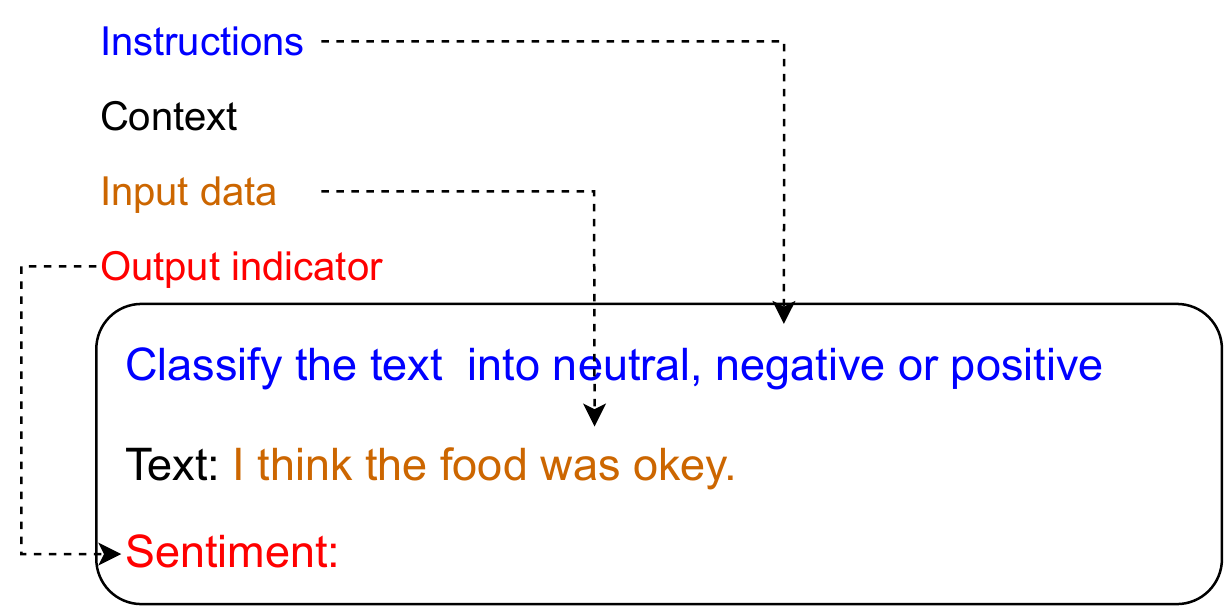}
    \caption{An example of a prompt.}
    \label{fig:prompt}
\end{figure}

% 6. Prompt engineering
\paragraph{Prompt-based and in-context learning}
Prompt-based learning is a relatively new paradigm that allows AI engineers and researchers to better understand the behavior of LLMs \citep{Liu-Prompting-survey-ACM-23}. Instead of estimating the conditional probability $P(y|x)$, prompt-based learning models the probability of text directly based on text generation. The input sequence $x$ is converted by using a \textit{template} to create a \textit{prompt} $x'$ that includes some unfilled slots. This prompt is put into LLMs for the final prediction of unfilled slots $\hat{x}$. The task of creating prompts is called prompt engineering.\footnote{https://github.com/dair-ai/Prompt-Engineering-Guide} Figure \ref{fig:prompt} shows an example of a prompt. The prompt includes three parts: instructions, input data, and output indicator. The context is not required in this classification task.

In-context learning refers to a learning paradigm in which a ML model acquires knowledge by processing and analyzing data within its relevant context, allowing it to draw meaningful insights and adapt its understanding based on surrounding information \citep{Liu-In-Context-GPT3-21,zhang2023makes}. This approach contrasts with traditional supervised learning, where ML models are trained on isolated instances or labeled data. In-context learning enables ML models to better generalize and adapt to new situations, improving their performance on complex tasks, and facilitating more accurate and nuanced responses in real-world applications.

LLMs work well with zero-shot and few-shot learning by using prompt-based and in-context learning \citep{radford2018improving,Schick-Cloze-Question-Few,brown2020language,Liu-In-Context-GPT3-21}. There are several prompt engineering techniques such as chain-of-thought (COT) \citep{wei2022chain}, zero-shot COT \citep{kojima2022large}, self-consistency \citep{wang2022self}, generated knowledge prompting \citep{liu2022generated}, and automatic prompt engineer \citep{zhou2022large} that can be used to distill and adapt the knowledge of LLMs for practical problems.

%Followed by a discussion on the factors that led to the breakthrough of LLMs
\subsection{Factors behind the Breakthrough of LLMs}
%This section shows the factors behind the breakthrough of LLMs.

% 1. Transformers and self attention
\paragraph{Seq2seq, attention, and Transformer}
The success of LLMs is technically derived from the emergence of seq2seq \citep{sutskever2014sequence}, attention \citep{bahdanau2015neural}, and Transformer \citep{Vaswani-attention-NISP-17}. The seq2seq provides a method that can naturally represent text-to-text problems by using the encoder-decoder architecture. The attention mechanism appears to support the seq2seq for better using the input via a contextual attention vector that is used for decoding. Transformer takes advantage of seq2seq and attention by designing a new encoder-decoder architecture with new components: self-attention, multi-head attention, and residual connections. The new architecture leverages the significant growth of PLMs such as BERT \citep{DCLT-NAACL-19} and LLMs such as GPT-3 \citep{brown2020language}.

\paragraph{Massive data}
% 2. Massive unlabeled and instruction data and a lot of annotated data for down-stream tasks
The second factor that leverages the significant growth of LLMs is the availability of massive unlabeled and annotated data of down-stream tasks. For example, GPT-3 (175B parameters) \citep{brown2020language} was pretrained with the CommonCrawl corpus\footnote{https://commoncrawl.org/the-data/} \citep{Raffel2020ExploringTL}. After pretraining, it works well on many NLP tasks \citep{brown2020language}. While common massive unlabeled corpora are necessary for training LLMs by using self-training, annotated data provides good information sources for task-adaptive pre-training of downstream tasks \citep{Gururangan-Dont-stop-pre-training-ACL-20}.

\paragraph{Self-training}
% 3. Self-training: Masked LMs, next sentence prediction,...
Besides the availability of massive corpora, self-training (self-supervised) is a critical concept for training LLMs. Different from the traditional ML mechanism that requires annotated samples created by humans for training, self-training augments unlabelled data to create training samples. The main idea of self-training is to automatically corrupt input sequences (i.e., masking some tokens of an input) and the learning process forces the model to reconstruct the original sequences based on the corrupted sequences. Self-training allows us for training LLMs without using annotated data. There are several self-training methods as follows \citep{Amatriain-Transformer-23}.
\begin{itemize}
    \item \textbf{Language model (LM)} is the traditional self-training technique that relies on the concept of language modeling. Given an input sequence, LM predicts next tokens given the previous tokens. The prediction includes unidirectional and bidirectional LMs. The unidirectional LMs only predict next tokens of the given input tokens in a uni-direction (forward or backward) while bidirectional LMs combine both forward or backward LMs for prediction.

    \item \textbf{Masked language model (MLM)} restores masked tokens (corrupted tokens) by using its surrounding context. Given an input sequence, the MLM first randomly masks some tokens and the training process forces the model to restore the masked tokens by comparing predicted tokens and gold tokens of the input sequence. For example, BERT \citep{DCLT-NAACL-19} randomly masks 15\% of all tokens for training the MLM. 

    \item \textbf{Permuted language model (PLM)} is the same as LM but uses a random permutation of input sequences.

    \item \textbf{Denoising autoencoder (DAE)} operates as the autoencoder mechanism. It takes a part of corrupted input sequences (randomly masking tokens or suffering input sequences in random order) and recovers the original unsorted input.

    \item \textbf{Contrastive learning (CTL)} is to learn a scoring function for text pairs of similar and random sampled sequences. It includes four different types: deep infomax, replaced token detection, next sentence prediction, and sentence order prediction.

\end{itemize}

\begin{figure}[!t]
    \centering
    \includegraphics[width=0.8\textwidth]{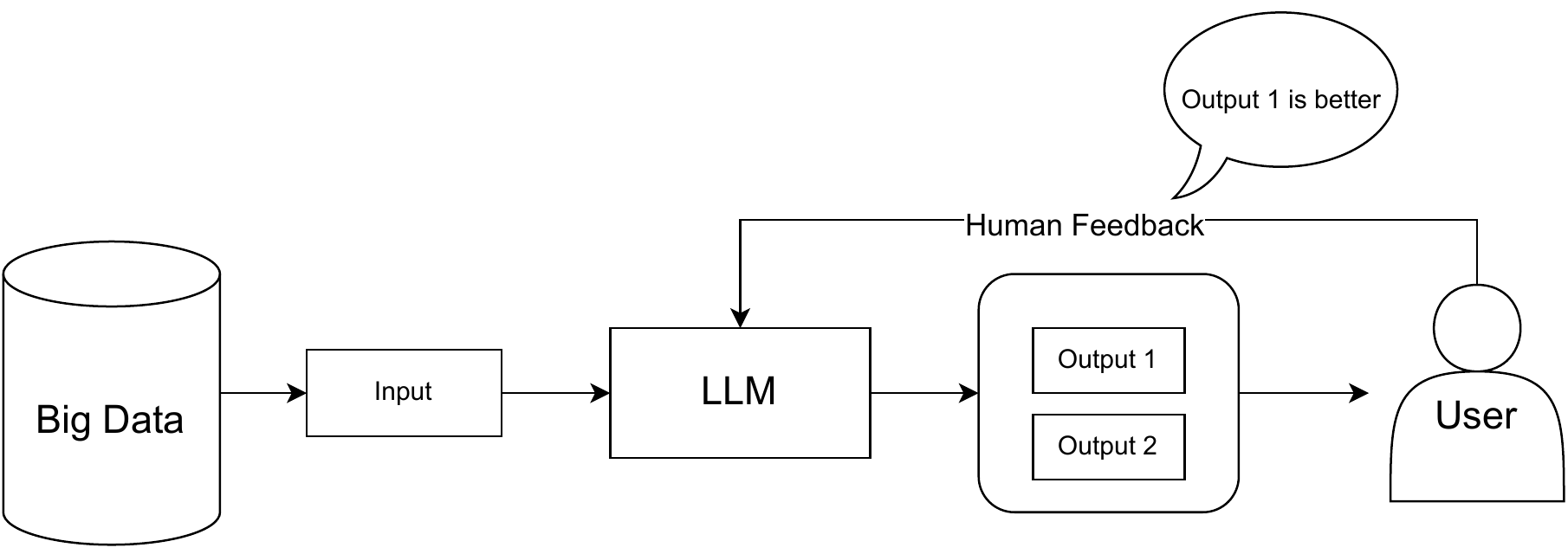}
    \caption{Training language models with human feedback. In this framework, the models are trained using human feedback in the form of preference.}
    \label{fig:hitl}
\end{figure}

\paragraph{Fine-tuning and human-in-the-loop}
Despite the significant advancements in early LLMs such as GPT-2 \citep{radford2018improving}, the self-training process still faces certain challenges. The issue of self-training is that LLMs only have access to unstructured data, and their objective is simply predicting the next tokens, given certain text. This approach has a problem of no intention, where the model generates content to make it looks like the training data, without understanding the context or the purpose behind it, which may lead to unaligned results with users' needs. 

Incorporating human feedback can enhance the quality of the generated content and ensure that it aligns with the desired outcome. The feedback provided by humans can be used to train the model more effectively and improve its overall performance. One commonly used form of feedback is user preference \citep{ziegler2019fine,nguyen2022make}. It is cheaper to evaluate the model's output using preference than labelling or a ranking system. In particular, given an input, LLMs will generate two output candidates and present them to a human evaluator, who will decide which one is better. The LLMs will utilize this preference feedback to refine its outputs based on the evaluator's preferences. Please refer to Figure \ref{fig:hitl} for more detailed information. 

\paragraph{Reinforcement learning with human feedback} 
Reinforcement learning with human feedback (RLHF) is an innovative approach that integrates human guidance into the reinforcement learning process to improve the performance of an agent in various tasks. By leveraging human expertise, the RLHF method can help to overcome limitations in designing reward functions, exploring new strategies, and facilitating faster convergence to optimal solutions \citep{christiano2017deep}

In a typical RLHF setting, the learning agent interacts with an environment and receives evaluative feedback from a human expert. This feedback is utilized to create a reward function that guides the agent's learning process. By incorporating human knowledge into the reward function, RLHF allows the agent to better understand the underlying objectives and constraints, leading to more efficient and effective learning.
By incorporating RLHF, LLMs can learn to generate more accurate and relevant content to maximize the feedback rewards. As testified in an Open AI's paper \citep{ouyang2022training}, the incorporation of human feedback in the form of preference (i.e., humans choose which output is better) increases the performance of LLMs consistently by about 50\% across different model sizes. 

% To sum up, the integration of human feedback and reinforcement learning can significantly enhance the capabilities of Large Language Models. By incorporating human in the loop, LLMs can generate content that is aligned with the desired outcome, while RL can enable LLMs to learn efficiently and effectively from human feedback. These advancements pave the way for the development of more advanced and efficient LLMs in the future such as GPT-3 and Chat-GPT.

\paragraph{The power of computing resources}
% 4. The power of computing resources
The power of computing resources (GPUs and TPUs) significantly contributes to the breakthrough of LLMs. GPUs and TPUs allow us to speed up the training of LLMs in a practical manner \citep{hoffmann2022empirical}, which significantly increases the size of pre-trained models from million to billion parameters.

\subsection{LLMs' timeline}
\begin{figure}[!t]
    \centering
    \includegraphics[width=1\textwidth]{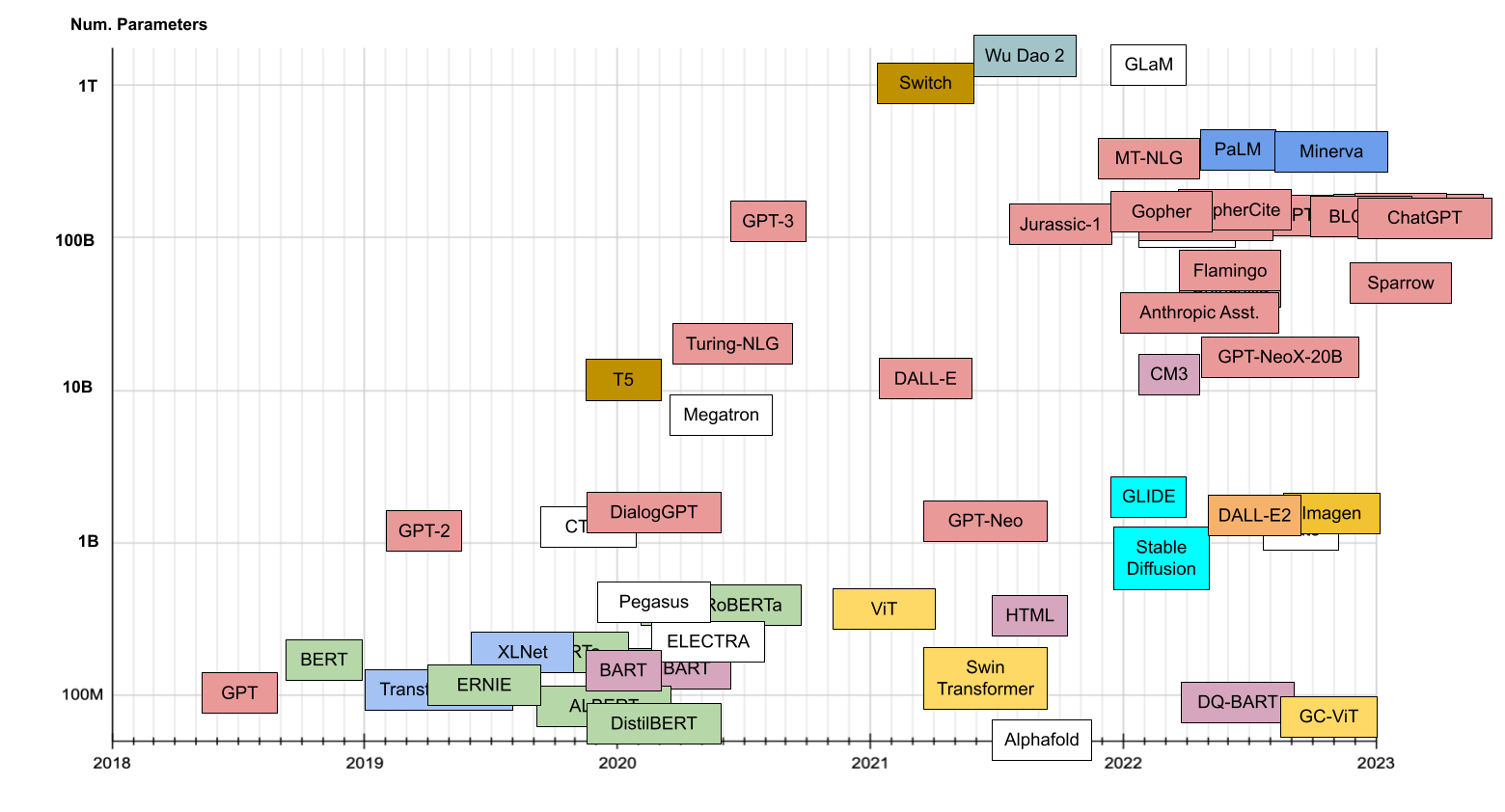}
    \caption{The timeline of LLMs with the number of parameters extracted from \cite{Amatriain-Transformer-23}.}
    \label{fig:llms}
\end{figure}

Figure \ref{fig:llms} shows LLMs' timeline and the number of parameters extracted from \cite{Amatriain-Transformer-23}. It is clear to observe the evolution of LLMs from 2018 to 2023. Started by small PLMs such as GPT and BERT in 2018, training LLMs has received significant attention of research and investment from both industry and academia \citep{bowman2023eight}. For more detailed information of LLMs and its catalog, we encourage audiences to read the paper of \cite{Amatriain-Transformer-23}.

The timeline of LLMs pushes the paradigm shift of NLP in which all NLP tasks can be represented as text generation \citep{radford2018improving,brown2020language,Gururangan-Dont-stop-pre-training-ACL-20,Du-All-NLP-arXiv-21}. For example, GPT-2 \citep{radford2018improving} showed that a generative pretraining LM improves the performance of 9 out of 12 NLP tasks, i.e., 8.9\% on commonsense reasoning (Stories Cloze Test), 5.7\% on question answering (RACE), and 1.5\% on textual entailment (MultiNLI). GPT-3 with few-shot learning \citep{brown2020language} is even better on the SuperGLUE benchmark \citep{wang2019superglue} that includes 8 challenging natural language understanding tasks: question answering, textual entailment, coreference resolution, word sense disambiguation, and causal reasoning. \cite{Du-All-NLP-arXiv-21} introduced a general pretraining framework that formulates NLP tasks as text generation. The framework also works well on the SuperGLUE benchmark. LLMs have continuously improved in terms of quantity and quality. For example, ChatGPT has more than 100 billion parameters and GPT-4 \citep{openai-gpt4} with even a bigger model trained by multimodal data can pass bar exams in the top 10\%. \\

% The purpose of this analysis
\textbf{The purpose of this paper} is to empirically analyze the behavior of LLMs in actual business cases. The main research question is ``\textbf{\textit{Do LLMs need knowledge enhancement of specific domains for working well in business scenarios}?}". To answer the question, we first show the applications of LLMs in Section \ref{sec:applications} and then discuss the challenges of LLMs in Section \ref{sec:limitations}. Based on the challenges, we introduce our proposal that enhances the domain knowledge of LLMs in Section \ref{sec:method} and describe experimental settings in Section \ref{sec:setting}. We finally report preliminary results of the question answering task in the insurance domain in Section \ref{sec:result} and draw conclusions and future directions in Section \ref{sec:conclusion}.

\section{LLMs' Applications}\label{sec:applications}
%LLMs have adapted for many NLP problems.
%This section reports up-to-date applications of LLMs.
We try our best to collect up-to-date applications built by using LLMs.\footnote{The content was partially derived from https://huyenchip.com/2023/04/11/llm-engineering.html} While some actual applications are potential such as data annotation and generation, programming, and bug fixing, or tale-to-your-data, further research is still needed for boosting LLMs to business scenarios.

\paragraph{AI assistant and Chatbot}
have become one of the most popular consumer use cases for AI technologies. From scheduling appointments and making notes to responding to emails and helping with various tasks, AI assistants have become indispensable in our daily lives. Big-tech companies provide their own AI assistant and chatbot solutions. For example, Quora has even released their own app Poe, allowing users to chat with multiple LLMs, showcasing the increasing prevalence of AI assistants in the market \citep{li2023chatdoctor}. With the support of LLMs, AI assistants and chatbots can better understand user intents, correctly predict entities, remember the longer context, etc. All these help to make AI assistants and chatbots to be smarter and faster, and behave like humans.

\paragraph{Error correction, programming, and gaming}
No doubt that ChatGPT is a cutting-edge artificial intelligence language model that has an impressive ability to answer follow-up questions. However, ChatGPT and LLMs have unique strengths that make them as valuable tools for tasks that require more nuanced language understanding beyond simple error correction. For example, ChatGPT has adapted for several tasks such as grammatical error correction \citep{wu2023chatgpt}, generating and debugging codes\footnote{https://www.youtube.com/watch?v=8y7GRYaYYQg} of different programming languages, debugging written codes, completing missed parts of codes, rewriting codes in other languages \citep{kashefi2023chatgpt}, solving mathematical problems, or creating realistic conversations with game characters.\footnote{https://www.youtube.com/watch?v=-R4PWIkgOSk}

%\cite{wu2023chatgpt} showed that while ChatGPT may not perform as well as other models in terms of automatic evaluation metrics for grammatical error correction, it has unique strengths in its ability to go beyond one-by-one corrections.
%\cite{kashefi2023chatgpt} examined the ChatGPT ability to generate codes for different programming languages, debug written codes, complete missed parts of codes, rewrite codes in other languages, and parallelize serial codes. Mathematical problems such as the Poisson equation, diffusion equation, Navier-Stokes equations, and eigenvalue problems are considered, along with scientific machine learning.

%ChatGPT can also used for gaming. It can generates codes for gaming programming,\footnote{https://www.youtube.com/watch?v=8y7GRYaYYQg} generate game characters, 

%While ChatGPT can successfully program numerical algorithms, certain limitations and challenges exist that require further improvement of the machine learning model.

%\cite{wu2023chatgpt} investigated ChatGPT on the Grammatical Error Correction (GEC) task. They showed that while it 

%This suggests that ChatGPT may be better suited for tasks that require more nuanced language understanding beyond simple error correction. For example, ChatGPT's ability to change surface expressions and sentence structure while maintaining grammatical correctness makes it an ideal tool for tasks such as content creation, where the goal is not just to correct errors but also to create engaging and compelling content. 

\paragraph{Data annotation and generation}
Another potential use case for LLMs and ChatGPT is text annotation commonly used in NLP applications. The study of \cite{gilardi2023chatgpt} found that ChatGPT outperforms crowd-workers and trained annotators for several annotation tasks on 2,382 tweets. Additionally, ChatGPT's intercoder agreement exceeds both crowd-workers and trained annotators for all tasks, and it is twenty times cheaper than MTurk. However, ChatGPT's performance across multiple languages and types of text, implementing few-shot learning, semi-automated data labeling systems, and improving zero-shot reasoning is still an open question.

%further research is needed to explore ChatGPT's performance across multiple languages and types of text, implementing few-shot learning, semi-automated data labeling systems, and improving zero-shot reasoning.

\paragraph{Learning and education}
%ChatGPT has gained popularity among students for its ability to assist with various tasks in education. From generating quizzes to providing feedback on essays, ChatGPT has been used in a wide range of applications to support learning. However, despite its capabilities, ChatGPT may face downtime or limitations in certain scenarios. In such cases, other AI-powered technologies can be leveraged alongside ChatGPT to further enhance its functionalities and provide more comprehensive assistance to students \citep{bull2023generative}.

For learning and education, LLMs and ChatGPT can support learners and teachers in many aspects. For example, LLMs can help to summarizing books or reports, automatically generating quizzes to test students' comprehension, grading essays and providing feedback, explaining math solutions \citep{agrawal2022mathematics}. This can help to save learners and teachers time for learning and teaching \citep{edii}. In addition to learning tasks, LLMs can also serve as debate partners. It can present arguments from different perspectives on a given topic, allowing students to practice critical thinking and persuasive skills by engaging in debates with ChatGPT. This can be a fun and interactive way for students to develop their argumentation and reasoning abilities.

\paragraph{Talk-to-your-data}
Talk-to-your-data is currently one of the most popular enterprise applications. Startups are continuously building tools to allow enterprise users to query their internal data and policies in natural languages or in a Q\&A format. Many of these startups focus on specific verticals such as legal contracts, resumes, financial data, or customer support \citep{saeed2023querying}.

%By utilizing a company's documentation, policies, and FAQs, chatbots can be developed to respond to customer support requests \citep{saeed2023querying}.

%Typically, this application involves four main steps. First, internal data is organized into a database, such as an SQL database, graph database, embedding/vector database, or text database. Second, given an input in natural language, it is converted into the query language of the internal database. For example, this could involve returning a SQL query for an SQL or graph database or an ANN (approximate nearest neighbour) retrieval query for an embedding database \citep{saeed2023querying}. For a purely text-based database, a search query may be extracted. Third, the query is executed in the database to obtain the query result. Finally, the query result is translated into natural language.

%While this application may have some impressive demonstrations, the defensibility of the category is unclear \citep{majmudar2022differentially} \citep{huang2020unexpected}. There are already startups building applications to allow users to query on top of databases like Google Drive or Notion, which can potentially be implemented as a feature within a week.

%OpenAI has developed a comprehensive tutorial on how to communicate with your vector database. While GPT-3.5-Turbo can detect patterns within small data that can fit into the input prompt, it may not be suitable for large production data analysis.

\paragraph{Search and recommendation}
LLMs have revolutionized the traditional keyword-based search approach.
%The application of Language Model Models (LLMs) in search and recommendation for enterprise use cases has revolutionized the traditional keyword-based search approach. LLMs enable a more contextual and personalized search experience, allowing users to find relevant information and recommendations even when they may not know exactly what they are looking for.
We have seen recently innovative approaches like Chat-Rec \citep{gao2023chatrec}, which combines LLMs with recommender systems to enhance interactivity and explainability.
%Chat-Rec converts user profiles and historical interactions into prompts for LLMs, improving the recommendation process and making it more interactive and explainable. The approach is effective in learning user preferences and establishing connections between users and products, and can handle cold-start scenarios with new items. 
%The experiments show that Chat-Rec improves the results of top-k recommendations and performs better in zero-shot rating prediction tasks. Overall, Chat-Rec presents a novel approach to improving recommender systems and opens up new practical scenarios for implementing AI-generated content in recommender system studies.
With some modification of GPT models, it is possible to build an app to be used as cross-encoders or bi-encoders for semantic search \citep{muennighoff2022sgpt}. 

\paragraph{Sales}
LLMs have numerous applications in the field of business management. LLMs can help with data querying, visualization (charts and graphs), or storytelling, where users can ask a question in a natural manner and receive a narrative around the insight associated with the users' questions. Another use case is to see LLMs for sales activities such as writing sales emails. In this case, LLMs can be considered as a powerful tool for reducing the effort of salesman.

\paragraph{SEO}
AI has revolutionized the field of search engine optimization (SEO) in several ways, as it enables website owners to focus on creating high-quality content and better services. By providing more efficient data-driven insights and automating workflow \citep{cao2023comprehensive}, LLM-powered apps can analyze search queries and suggest relevant keywords to users, making it easier to identify patterns and trends.
%These apps also optimize the length, structure, and readability of content and suggest relevant keywords with much broader context in internet, making websites more SEO-friendly.
In addition, LMMs can help to generate data as much as SEO companies want to increase the rank of webpages processed by ranking algorithms.

%For optimal performance and handling complex tasks, models must continuously improve and optimize themselves. User feedback plays a crucial role in this optimization algorithm. LLMs can engage users in feedback loops to collect user responses and use them to optimize the model. The process involves presenting the model's prediction results to the user and requesting feedback \citep{nguyen2022make} , which can be direct or free-form. Once enough feedback data is collected, the model can analyze it using natural language processing techniques and machine learning algorithms to adjust and optimize the model \citep{cao2023comprehensive}.

\section{The Challenges of LLMs}\label{sec:limitations}

% Emphasize LLMs' limitations. Listing all limitations of LLMs and try to pick up 1-2 limitations that we can address in the next sections. 

Productizing LLMs is no easy feat. Various challenges must be overcome for LLMs to achieve commercial viability, including tackling engineering and technical difficulties, addressing limitations in capabilities, and preventing ethical or social problems that may arise from their usage. It makes challenges to integrate LLMs into existing products or create new ones.
%Overcoming these challenges can be arduous, making it challenging to integrate LMs into existing products or create new ones. 

\subsection{Engineering Obstacles}

One of the biggest development roadblocks is scalability \citep{tamkin2021understanding}. The training of LLMs is a computationally intensive task, requiring massive amounts of data and heavy computing power. Hence, scaling an LLM to handle business-specific tasks with large data can be daunting. Most of the time, LLMs are provided by third-party services, which cost fees and complicate large-scale product customization.  

%As mentioned earlier,
An additional hurdle is ensuring data confidentiality and protection. LLMs are often deployed on the cloud hosted by third-party companies. So there is no guarantee that the data sent to LLMs is untouched. It is dangerous to applications using sensitive data, such as personal conversations, medical records, and financial transactions. 
% It is crucial to ensure that this data is confidential and not misused by third-party agents or hackers.
There are solutions to mitigate the issue, but the performance is unsatisfactory and comes with additional development costs \citep{yudifferentially,lilarge}.

There are also challenges to integrating LLMs into existing systems. LLMs are often used with other technologies, such as chatbots or virtual assistants. Integrating these different technologies can be complex, requiring expertise in multiple domains such as natural language processing, machine learning, and software engineering. Therefore, it is a non-trivial task for ordinary people to use LLMs to address their business tasks.

\subsection{Capability Limitations}

\begin{figure}[!t]
    \centering
    \includegraphics[width=0.8\textwidth]{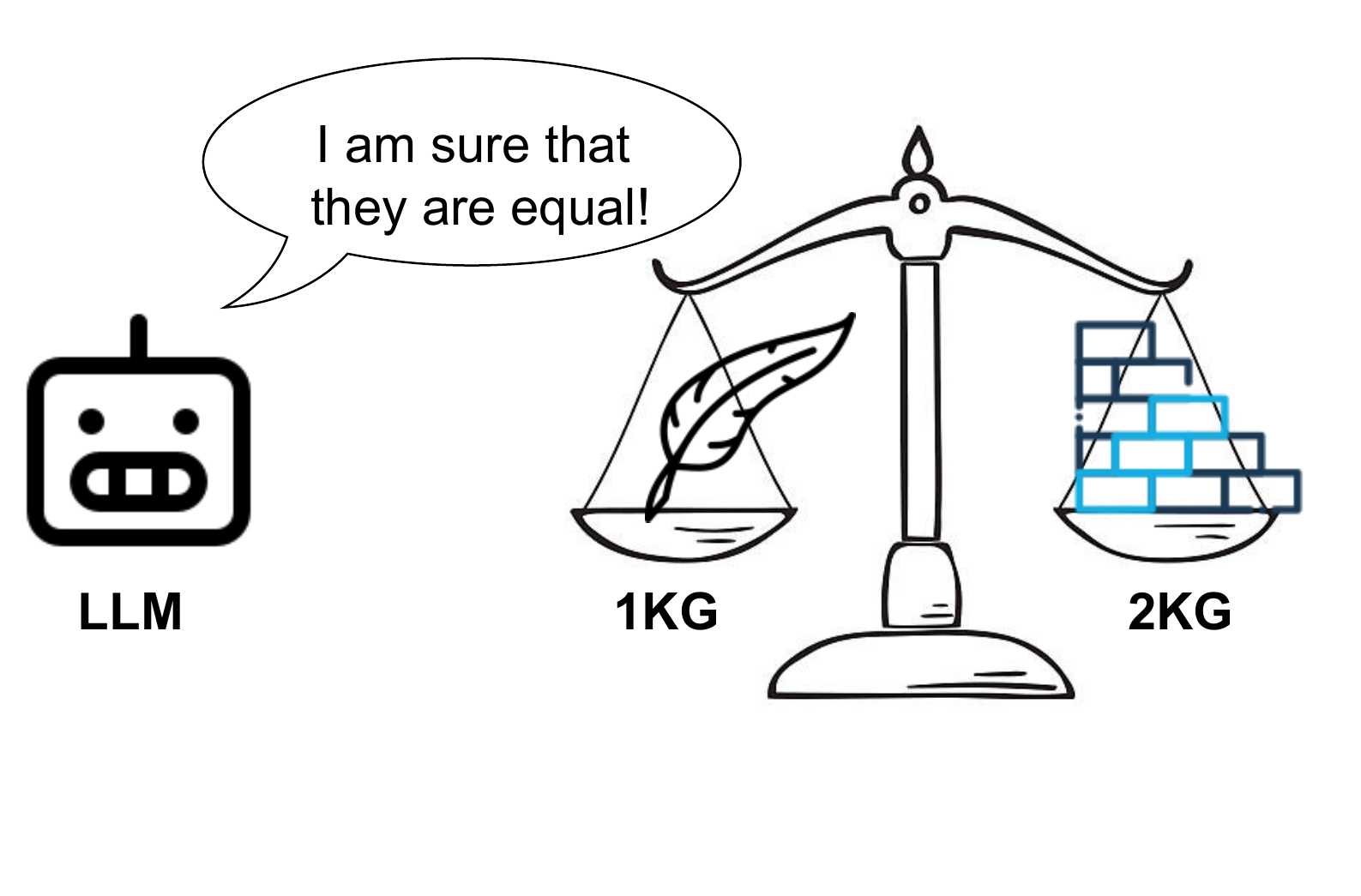}\vspace{-1.3cm}
    \caption{A simple question that can cheat LLMs: ``Which weighs more, two kilograms of bricks or a kilogram of feathers?"}
    \label{fig:2kg1kg}
\end{figure}

Although LLMs are powerful AI models, they are still far from the human level of intelligence. As illustrated in Figure \ref{fig:2kg1kg}, when answering the question \textit{``Which weighs more, two kilograms of bricks or a kilogram of feathers?"},  GPT-3 (version Mar-2023)  insists that both weigh equally because it is trained with the common questions comparing one-kilogram bricks versus one-kilogram feathers. Worse still, the model would try to convince it was right by providing a complicated, yet unrelated explanation: \textit{``Two kilograms of bricks and one kilogram of feathers weigh the same amount, which is two kilograms or 2000 grams. ... So, even though two kilograms of bricks consist of fewer individual pieces than one kilogram of feathers, both quantities have the same mass and, therefore, the same weight."}. 

The above example illustrates LLMs' tendency to produce hallucinations \citep{azamfirei2023large}. These hallucinations occur when LLMs generate text that is not grounded in reality which is a challenge to distinguish the real and the imagination. In real-world settings such as insurance businesses for example; a chatbot trained on policy documents may answer users' questions by providing fake content about policy packages or claim conditions that appear genuine, which is confusing, and potentially leads to harmful consequences. 

Another significant technical limitation is the reasoning capability of LLMs \citep{Bang-ChatGPT-Evaluation-23}. While they excel at pattern recognition and generating text based on statistical patterns, they may lack the ability to reason or understand concepts beyond the surface level. This limits their ability to understand complex contexts or make logical inferences, which can be crucial for many applications. For instance, a model that generates legal documents must be able to reason about legal precedent and apply it to specific cases.

% Similarly, reading through terms and conditions in insurance rule books often requires sophisticated reasoning to combine terms from distant sections. 

LLMs exhibit extremely low interpretability, posing a significant challenge for building trust in their applications. LLMs are like black boxes, making it difficult to understand how they predict or make decisions. This lack of transparency can be a problem in fields such as healthcare, where decisions made by the model can have a significant impact on people's life. 

Finally, LLMs struggle with handling and generating structured data. While they can generate text, they struggle to produce structured data like tables or graphs, which can be important for many applications.
This limitation hinders their usefulness in tasks involving the processing of structured data such as tables, diagrams, equations, and figures.

\subsection{Ethical and Social Risks}

%In addition to the technical issues of LLMs, 
There is a rising ethical concern about the potential of LLMs to perpetuate harmful biases or even create threats to humans. These biases can be present in the data used to train the models, which can lead to unfair outcomes in decision-making processes. For example, if a language model of a chatbot is trained on data that is biased against certain racial or gender groups, it can generate biased dialogues that offend users, which dramatically deteriorates the quality of customer services. 

% Another concern is the potential for language models to be used for malicious purposes, such as spreading misinformation or propaganda. Since these models can generate convincing text, they can be used to create fake news articles or social media posts that can mislead people and cause harm to individuals or society as a whole.

Moreover, LLMs can also pose risks to individual privacy. This is a serious issue in business applications. For example, if an LLM is trained on personal client data, such as emails or social media messages, it can learn personal information about individuals and use it in unintended ways. Hackers can utilize this loophole to steal secret information from clients. This raises concerns about data privacy and the potential for misuse of personal information. Keen readers are referred to \citet{isaac2021ethical} for details of these ethical and social risks. 

% In addition to these ethical risks, there are also social risks associated with the use of language models. For example, language models can lead to the loss of jobs for people whose work involves writing or content creation. As language models become more advanced, they may replace human writers altogether, which could have significant implications for the job market. 

% Data out-date

% Correctness

% Hallucination

% Over-confidence

\section{LLMs$^+$: Knowledge Enhancement for Domain Adaptation}\label{sec:method}
This section first describes a business scenario that requires the reasoning ability of LLMs and shows the proposal for domain-specific knowledge enhancement that adapts LLMs to actual cases.

\subsection{A business scenario: Question Answering for Insurance}
As mentioned in Section \ref{sec:applications} LLMs can be adapted for many applications. To provide a better understanding of LLMs in actual business cases, we adapt LLMs for insurance because of two main reasons. First, insurance is the traditional business with a big market\footnote{https://www.oecd.org/pensions/globalinsurancemarkettrends.htm} \citep{benfield2015global} where many insurance companies provide their packets for clients. If AI solutions can be successfully applied to (partly) automate insurance processes, it can boost the growth of the market and significantly reduce the cost of labor-expensive tasks. Second, insurance activities usually require a deep understanding of complex policies, regulations, and legal jargon for insurance processes. These activities are much more challenging than standalone tasks such as information extraction or information retrieval which AI models can achieve promising results. We leave the analysis of LLMs for other domains such as IncidentAI or LegalAI as a future task, after having preliminary results of LLMs for insurance.
While there are many domain-oriented tasks of insurance, the question-answering (QA) task was selected as a case study for the analysis due to two reasons. First, QA is a common task of insurance companies who have to handle many requests from customers per day for asking information of relevant procedures, e.g., claim making or the advice of insurance products, etc \citep{nuruzzaman2020intellibot}. A QA system comes in as a powerful tool that can quickly and accurately answer complicated insurance-related questions. Second, QA is a challenging NLP task that requires a deep understanding of the meaning of questions and context, and the implicit mutual relationship between them for giving correct answers \citep{rajpurkar2016squad,baradaran2022survey}.

In practice, end users usually ask questions about their products of insurance by making a phone call or by using a chat interface. For example, the question \textit{``Is there any refund for cancellation?"} is handled by insurance staff who will first confirm the personal information of users. After confirmation, the staff will retrieve relevant information in policy rulebooks or in-house rules to answer the question. The answer to the question is \textit{``Yes"} with references extracted from the rulebooks. Please refer to Table \ref{tab:i_gpt_output_sample} for more examples. However, the large amount of human resources for this QA process is the pain-point of insurance companies due to two reasons. First, they have to maintain a lot of people for this QA task. Second, in many cases customers have to wait a quite long time for an answer and human errors can happen in the operation process. Insurance companies have spent a lot of effort in building customer care centers, so they want to improve customers' experience by making collaboration between humans and AI. The use of AI in insurance-related tasks is so beneficial in terms of response time, correctness, and cost saving.

\subsection{The proposed model}\label{sec:model}
% show our proposal that combines LLMs with domain experts, KG, etc. LLMs are just an agent in the whole AI process

\paragraph{Problem statement}
As mentioned, LLMs have achieved promising results for many NLP tasks \citep{radford2018improving,Schick-Cloze-Question-Few,brown2020language,Liu-In-Context-GPT3-21}. However, we argue that LLMs need knowledge enhancement for domain adaptation by using an existing enterprise knowledge base. It comes from the challenges of LLMs mentioned in Section \ref{sec:limitations}. First, LLMs are trained by using a massive amount of data with billions of parameters. It allows LLMs can work well in common cases (Section \ref{sec:applications}). However, it may lack domain knowledge of specific downstream tasks. For example, if a question includes a specific information, e.g., the maximum amount of advanced medical care benefits, LLMs alone can only response general information (Section \ref{sec:output-observation}). As the result, LLMs need to connect and process policy rule books, in-house rules of insurance, or external knowledge, e.g., DBpedia when answering questions about claims from users. Second, the reasoning capability of LLMs is still an open question. Though the reasoning capability can be significantly improved by using prompt-based learning techniques \citep{wei2022chain,kojima2022large,wang2022self,liu2022generated,zhou2022large}, the lack of domain knowledge still makes a challenge for LLMs in making correct prediction. For example, given the question of \textit{``How much is the radiation treatment benefit payment?"}, GPT can only answer with general information while combining GPT with insurance policy rulebooks, it can reason the radiation treatment is 10 times of the daily amount of the hospitalization benefit (see Table \ref{tab:i_gpt_output_sample}).\vspace{-0.2cm}

\begin{figure}[!h]
    \centering
    \includegraphics[width=0.8\textwidth]{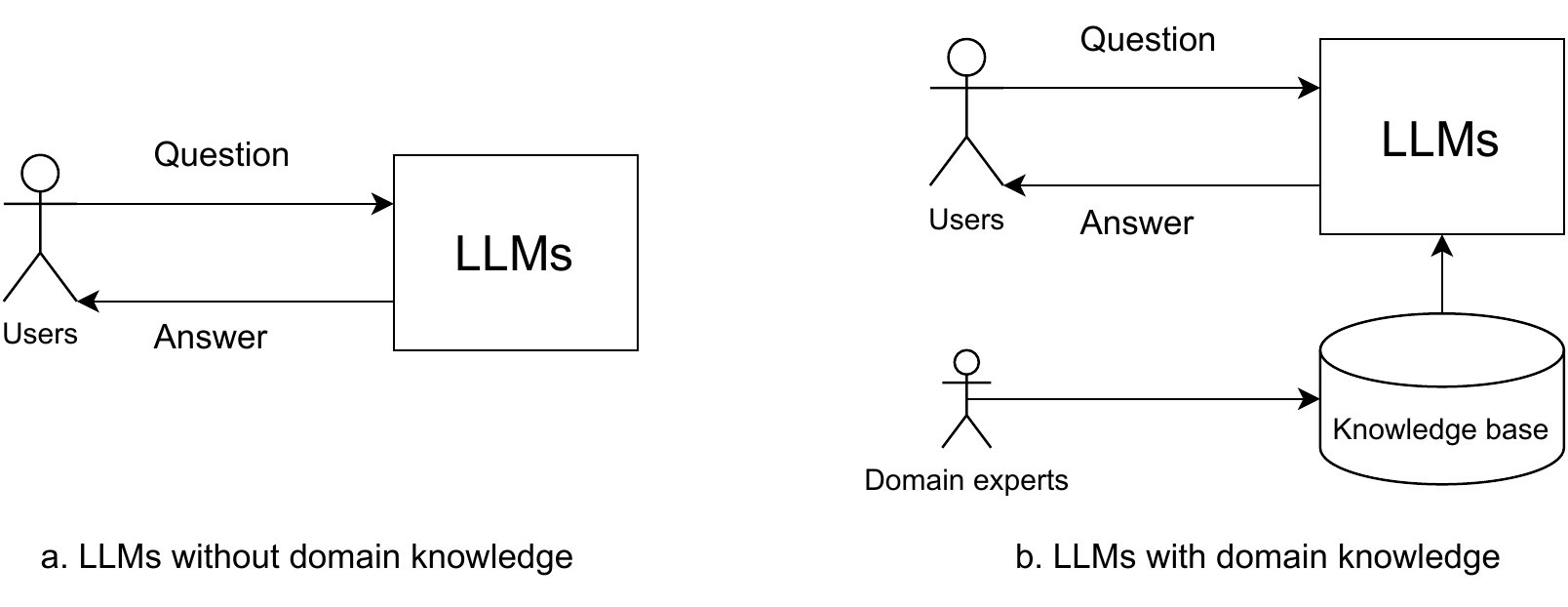}
    \caption{The proposal of LLMs using a domain knowledge base.}\vspace{-0.2cm}
    \label{fig:model}
\end{figure}

\paragraph{The model}
We address the lack of domain-specific knowledge of LLMs by introducing a new architecture in Figure \ref{fig:model}. The left hand side is the general process of using LLMs for down-stream tasks by using zero-shot or few-shot learning with prompting techniques. Users ask questions by providing prompts and LLMs response appropriate answers. The right hand side, in contrast, shows our proposal that adapts LLMs for domain adaptation. The knowledge of LLMs is enhanced by using a domain knowledge base created by domain experts. For example, if users ask a question about the obesity of insurance contracts, LLMs have a little chance to correctly answer this question due to the lack of domain knowledge. By leveraging LLMs with domain-specific knowledge, the system can understand and respond to questions in a conversational manner, making it easy for both insurance agents and customers for interactions. While any LLMs can be used in the proposed model, we take GPT-3.5-Turbo\footnote{We utilized both GPT-3.5-Turbo as well as text-davinci-003 and got the same performance, while the cost of the former one was as 10 times as lower than the later one. We are testing with GPT-4 for the analysis and results will be available soon.} as the main backbone due to its quality for many NLP tasks.\footnote{https://platform.openai.com/docs/models/overview}
%The following section will provide more context and explanation of our experiment results to help you better understand the potential of LLMs for insurance.

\subsection{Domain Knowledge Implementation}
As mentioned, LLMs are enhanced by using domain knowledge. In this study, we used policy rulebooks and DBpedia as a type of external domain knowledge.
%More precisely, the domain knowledge includes two types: raw text from policy rulebooks and a knowledge graph (KG) created from the rulebooks.
It allows us to investigate the capability of LLMs to work with different data types.

% Hector, please help to write this part
\paragraph{Policy rulebooks}
play an important role in insurance. It covers all definitions, terms, policies, regulations, etc that are necessary for insurance-related tasks. Based on government laws, insurance companies define and provide their policy rulebooks for their in-house operation. A rulebook is a document that includes the purpose of the insurance, definitions for terms used in the document, general provision, payment of benefit, refund policies, and special conditions of the policies.

For data preparation, seven rulebooks published by an insurance company were first collected. Due to the limitation of LLMs for dealing with structured data, extracting text from PDF files is a crucial step for data preparation. To ensure the text is machine-readable and usable in the downstream QA task, preprocessing PDF files is necessary. The preprocessing step involves removing any non-text elements such as images using PDFMiner,\footnote{https://github.com/pdfminer} and parsing tables to the machine-readable format using Camelot\footnote{https://github.com/camelot-dev/camelot} (results of table processing is mentioned in Section \ref{sec:result}). For rulebooks, we only consider the raw text as LLM's input rather than the structure of the rulebooks. Once the PDFs have been preprocessed and converted to the text format, they can be used as input data for inference with LLMs, i.e., GPT-3.5-Turbo. In practice, the rulebooks were segmented into manageable chunks and each chunk was converted into a corresponding vector for information retrieval. By leveraging the textual content from policy rulebooks, LLMs can understand new concepts of insurance and then can appropriately answer questions from end users.

%generate high-quality text, answer questions, or perform other language tasks. This has wide-ranging applications, from improving customer service chatbots to automating document classification and analysis. Overall, extracting text from PDF files is an important step in harnessing the power of large language models for natural language processing tasks.

\begin{figure}[!h]
    \centering
    \includegraphics[width=0.7\textwidth]{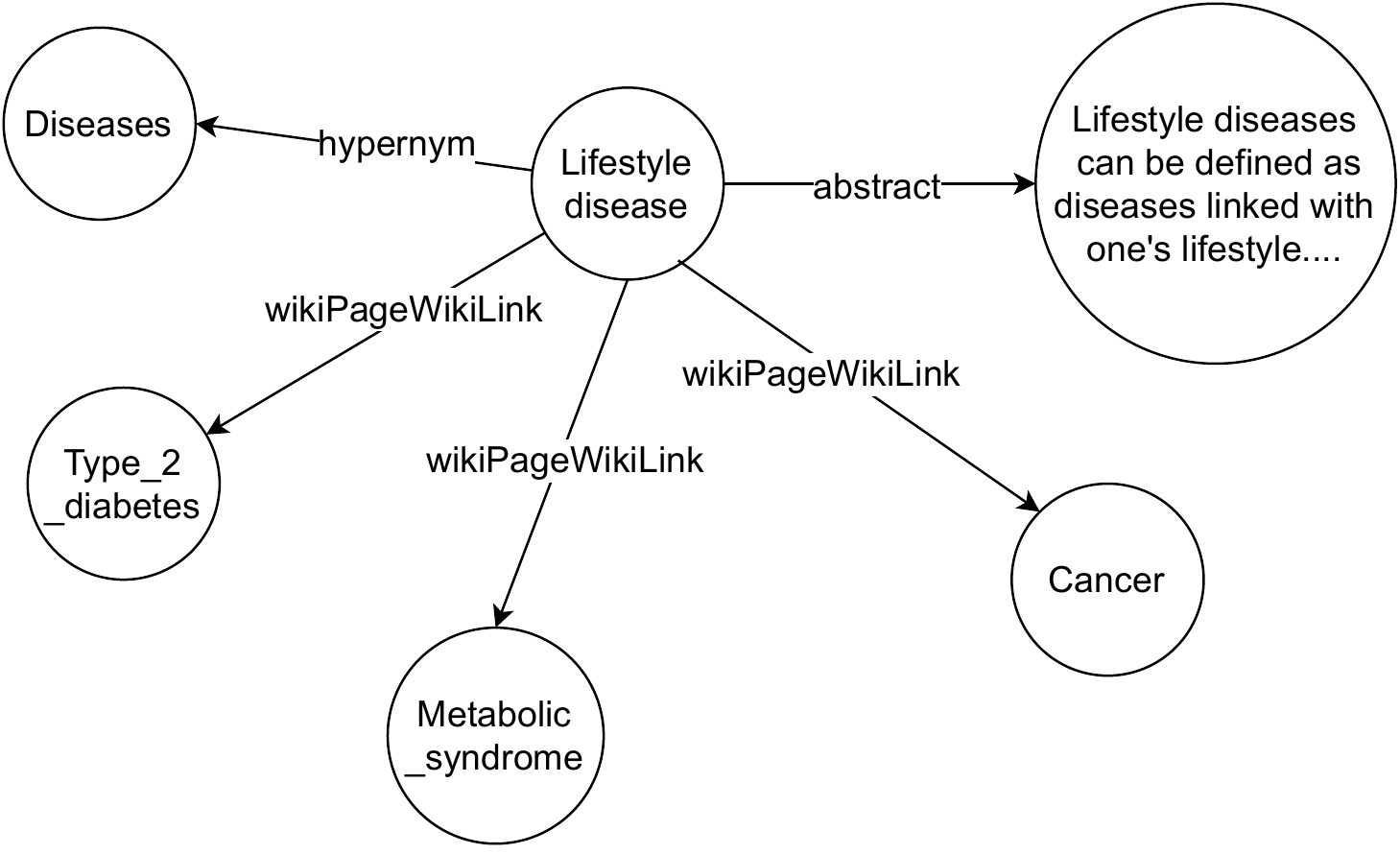}
    \caption{An example of the DBpedia KG.}
    \label{fig:dbpedia_sample}
\end{figure}

\paragraph{DBpedia as external knowledge}
Policy rulebooks provide good context for answering input questions, yet in practice, the questions may include concepts or entities that are not mentioned in the rulebooks.
For example, let's consider the question, \textit{``He was hospitalized for a week due to diabetes. How much is his allowance?"}. If GPT relies solely on the rulebooks, it may not understand that diabetes is a lifestyle disease.
In this case, external knowledge, e.g., DBPedia\footnote{https://dbpedia.org/} comes as an additional channel for enhancing information of the context extracted from rulebooks. For instance, the model can retrieve information such as, \textit{``Lifestyle diseases can be defined as diseases linked with one's lifestyle... which can lead to heart disease, stroke, obesity, type II diabetes, and lung cancer... heart disease, hypertension, metabolic syndrome..."}. An example of retrieving additional information from DBpedia is shown in Figure \ref{fig:dbpedia_sample}. By incorporating DBpedia's knowledge, the context of an input question is enhanced, resulting in more confident and accurate responses.

%Instead of using the domain-specific KG created from the policy rulebooks that do not include some lifestyle diseases, we used DBPedia\footnote{https://dbpedia.org/} as an external public KG. Since it contains verified facts, the chances of the algorithm generating erroneous information, i.e., hallucinations \citep{ji2023survey} can be minimized. To do that, given a question that contains lifestyle diseases, the model first retrieves relevant concepts of the diseases from DBPedia and uses these concepts as additional knowledge combined with the processed data from tables for reasoning. 

%However, with the integration of DBpedia retrieval, we can 

DBpedia follows the Resource Description Framework (RDF) format. RDF is a standard for expressing structured information on the web using subject-predicate-object triples, which is also known as the representation of a knowledge graph. In DBpedia, entities are represented as resources (subjects), their attributes are represented as predicates, and the attribute values are represented as objects. This RDF structure is flexible for querying and linking data. 
DBpedia offers localized versions in 125 languages, in which the English version contains 4.58 million entities, with 4.22 million entities classified in a consistent ontology.  
With the huge amount of facts and its structure, we can efficiently query, explore, and utilize structured data for various domains.

\subsection{Knowledge enhancement for QA}
\begin{figure}[!h]
    \centering
    \includegraphics[width=0.85\textwidth]{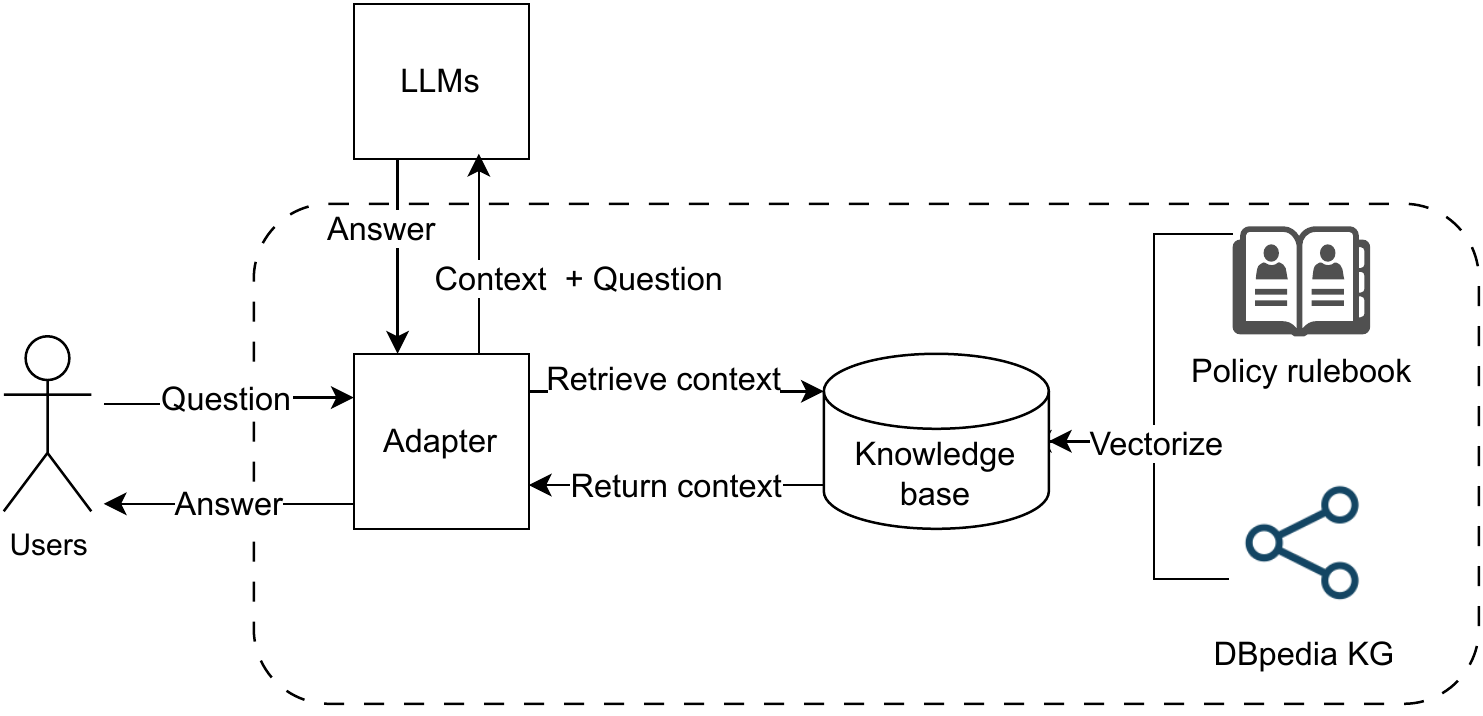}
    \caption{The proposed QA pipeline that utilizes domain-specific knowledge.}
    \label{fig:qa_pipe}
\end{figure}

% Hector
In NLP, incorporating external knowledge sources into LLMs has shown efficiency to improve the performance of QA models \citep{peng2023check}. In our experiments, we employed policy rulebooks and DBpedia to enhance knowledge of LLMs and define a QA pipeline in Figure \ref{fig:qa_pipe}. Instead of directly using the knowledge of LLMs, the pipeline first recieves a question by using an adapter that retrieves relevant context from the knowledge base created by the rulebooks. The retrieved context is combined with the input question to form a prompt processed by LLMs to generate the final answer.

%The methodology involves feeding the user's query and the relevant information from the rulebook or KG to a LLM such as GPT. This enables the LLM to leverage the latest context and provide prompt and precise information retrieval from the user's data sources. 

\paragraph{Information retrieval (IR)}
IR is an important module of the adapter that retrieves relevant context for an input question.
%This is because LLMs (such as GPT-3.5) can only process an input up to 4096 tokens,\footnote{The results of testing with GPT-4 will be available soon.} so adding all the rulebooks to the prompt is impossible.
For doing IR, raw chunks from rulebooks were first stored in a database in advanced. Given an input question, the adapter first converts it into a vector. For rulebooks, the IR computes the similarity of the question vector to vectors of all chunks. For DBpedia KG, the IR process begins by extracting entities from input questions using a method that combines part-of-speech tagging, named entity recognition, and a heuristic-based approach. These extracted entities act as candidate mentions for linking to the DBpedia KG. Then a text search engine\footnote{https://www.elastic.co/} is used to map these entities to corresponding entities within the DBpedia KG (an example of a part of DBpedia KG is shown in Figure \ref{fig:dbpedia_sample}). This mapping establishes connections between user queries and relevant entities within the KG. To retrieve specific information of these entities, the SPARQL (SPARQL Protocol and RDF Query Language) was employed. By formulating SPARQL queries, the model can interact with the DBpedia KG and retrieve desired information via the DBpedia SPARQL endpoint.
The retrieved chunks from the rulebooks and specific information of entities from DBpedia (Figure \ref{fig:dbpedia_sample}) were used as the context of an input question.

%This process ensures the extraction of precise and relevant data from the vast knowledge repository provided by DBpedia, facilitating efficient and accurate information retrieval for a given query.

% For rulebook-based KG, the IR estimates the similarity of the question vector to vectors of relevant text of triples. After computing, top $k$ relevant text (chunks or relevant text of triples) were selected as the context ($k=3$ for the rulebooks and $k=5$ for KG). The IR uses a simple Cosine similarity for relevant estimation.

\paragraph{Prompt creation}
Once the context has been retrieved, two types of prompts were designed for utilizing domain knowledge from the rulebooks and DBpedia. This is because the retrieved context from the rulebooks and DBpedia is different in terms of representation. The context directly retrieved from the rulebooks contains consecutive sequences while the context extracted from triples includes non-consecutive sequences (see Algorithm \ref{lst:prompt_rev_rulebook} and \ref{lst:prompt_rev_kg}). For example, the first prompt asks GPT-3.5 to answer a question of \textit{``When will I be notified of payment reason change?"} given the raw context.
The second prompt asks the question \textit{``He was hospitalized for a week due to diabetes. How much is his benefit?"} based on the context of triples. More precisely, a prompt includes four sections: (i) the indicator (\textit{``Answer the question in a short and concise way"}), (ii) the question, (iii) the context, and (iv) external knowledge. Detailed information of the two defined prompts can be found in Algorithm \ref{lst:prompt_rev_rulebook} and \ref{lst:prompt_rev_kg}.

%Nonetheless, a token limitation poses a challenge for LLMs such as GPT-3.5, which can only process input with a token count of up to 4096. As a result, it is not feasible to input the entire rulebook into GPT. To address this limitation, we employ a strategy that narrows down the scope of the rulebook. First, we present the rulebook in two formats: raw text and a knowledge graph. We then segment the rulebook into manageable chunks and generate corresponding embeddings for each chunk. When presented with a user's query, which is also embedded into a vector, we conduct a similarity search utilizing cosine similarity. In the experiment, we named these chunks are relevant rulebook section and relevant KG. Subsequently, we select the most similar chunk and utilize it as context when feeding it to GPT.

Once the prompts have been defined, it was applied to all the questions. All the prompts were iteratively fed into GPT by using its APIs to obtain corresponding answers.

\section{Experimental settings}\label{sec:setting}

\paragraph{Experimental scenarios}

%In technical terms, the approach entails leveraging a language model (LLM) to process insurance rulebooks and a user's natural language query, thereby generating natural language responses. To assess the efficacy of the methodology, we conducted a comprehensive evaluation of GPT-3.5's Insurance Question Answering capability. Specifically, we compared the performance of GPT-3.5 when using its internal knowledge as a knowledge base (KB) with that of using additional rulebooks as a supplementary KB.

We confirm the proposed model in two scenarios. The first is to directly use GPT-3.5-Turbo for answering questions. The second is to use GPT-3.5-Turbo with additional domain knowledge (context-aware) for QA. The total cost of experiments is around 250 USD.

 \begin{itemize}
    \item \textbf{Context-agnostic question answering}
 Given a user's question, we feed the question directly to GPT and use GPT's response as an answer. In this setting, GPT uses its own knowledge to answer input questions. The prompt of this setting is shown in Algorithm \ref{lst:prompt_simple} that contains two parts: (i) the indicator (\textit{``Answer the question in a short and concise way"}) and (ii) the question.
    
    \item \textbf{Context-aware question answering}
 Given a user's question, we feed the question and relevant external context to GPT and use GPT's response as an answer. This setting allows GPT to learn from context for knowledge enhancement. The context is information extracted from rulebooks or the combined information from rulebooks and DBPedia. The prompt of this setting is mentioned in Algorithm \ref{lst:prompt_rev_rulebook} and \ref{lst:prompt_rev_kg}.
 \end{itemize}

\paragraph{Data preparation}
Two datasets were used for performance evaluation.
%Seven PDF rulebooks and a collected dataset were first collected from an insurance company. 
%The rulebooks outline the terms and conditions of an insurance policy. It includes the details of the provided coverage, the premiums to be paid, and the conditions of a claim that can be made. The rulebooks typically contain information such as the policyholder's responsibilities, the insurer's obligations, and the exclusions and limitations of the policy. It also includes information on how to file a claim, the procedure for resolving disputes, and the consequences of non-compliance with the policy terms. 
The first is an expert-created dataset that consists of 104 gold question-answer pairs from an insurance company (Table \ref{tab:samples} shows some samples). Questions refer to different aspects of insurance policies and benefits. Some of them relate to refunds and surrender values in cases of cancellation, while others inquire about death payment benefits or hospitalization with specific medical conditions. There are also questions about the types of surgeries covered by the insurance policy and the amount of benefits payable for these procedures. Some questions mention the process of changing beneficiaries, payment of premiums, and the maximum payout limits for certain benefits. The answers to these questions depend on the specific insurance policy and the terms and conditions outlined in the contract.

The second corpus is a synthesized dataset.
This is because the expert-created dataset focuses on specific questions and answers related to insurance policies and benefits which are usually available in the rulebooks. However, in practical scenarios, questions may not always explicitly mention concepts or entities that are present in the rulebooks. To address this aspect, a synthesized dataset was created by combining commonsense knowledge from DBPedia and domain-specific knowledge from the rulebooks. This synthesized dataset captures a broader range of questions by generating queries based on smaller segments of the rulebooks and retrieving relevant information from DBPedia. The generated questions are then adjusted to incorporate both the rulebook information and the retrieved knowledge from DBPedia. More precisely, the synthesized dataset was created in the following steps.
First, the rulebooks were divided into smaller segments to ensure they adhere to the token limitations of GPT.
After chunking, GPT was employed to generate questions that pertain to each specific chunk.
Next, concepts (entities) in a generated question were extracted and used as queries for retrieving relevant information from DBPedia.
After that, the generated question was modified by asking GPT to adjust the question by considering both the chunk from the rulebooks and relevant information from DBPedia.
Finally, the GPT was requested to generate the corresponding answer for the synthesized question.
The creation process involves humans for manual checking to ensure the high quality of generated questions.
The resulting synthesized dataset contains 87 question-answer pairs after manual checking and removal of duplicates, providing a more comprehensive and diverse set of questions for evaluation purposes.

%Incorporating commonsense knowledge was a crucial aspect of the data synthesis process. To achieve this, we leveraged DBpedia, an automatically extracted knowledge base from Wikipedia, which contains a vast amount of general knowledge. We extracted relevant DBpedia data that was related to the synthesized questions. This data served as a source of commonsense knowledge. To further enhance the dataset, we tasked the GPT model with modifying the questions in a way that they would incorporate both the given rulebook chunk and the relevant DBpedia information. This process ensured that the new questions would require an understanding of both the specific insurance rulebook and general knowledge from DBpedia. This dataset encompasses a broader scope of insurance-related scenarios, enabling the model to provide nuanced and well-informed answers to a wide range of insurance-related questions.

\begin{table*}[t]
    \centering
    \caption{Samples of human questions. Top five questions with its outputs are showed in Table \ref{tab:i_gpt_output_sample}.}\label{tab:samples}
    \begin{tabular}{|l|}
    \toprule
    Question  \\
    \midrule   
    What kind of surgery does a hysterectomy involve? \\
    Is there a refund for cancellation? \\
    What is the maximum amount of advanced medical care benefits? \\
    When will I be notified of a payment reason change? \\
    How much is the radiation treatment benefit payment? \\
    What is an endowment insurance policy? \\ 
    When will I be notified of a payment reason change? \\ 
    I would like to know the payment amount of sickness hospitalization benefits \\
    How much is the advanced medical care lump-sum benefit? \\ 
    What kind of surgery does an oophorectomy involve? \\ 
    How much is the radiation treatment benefit payment? \\
    What kind of surgery does a hysterectomy involve? \\
    Is hospitalization for childbirth covered by hospitalization benefits? \\
    He was hospitalized for a week due to diabetes. How much is his allowance? \\
    \bottomrule
    \end{tabular}
\end{table*}
\paragraph{Evaluation metrics}
The evaluation was done by using humans instead of using both automatic and human evaluation. This is because the quality estimation of answers from the system requires a deep understanding of policy rulebooks, terms, and conditions relating to input questions rather than simply using token overlapping metrics such as ROUGE-scores \citep{lin2004rouge} or BLEU-scores \citep{papineni2002bleu}. Human evaluation was done in two steps. In the first step, two annotators who had experience with answering insurance questions were involved the evaluation process. Each annotator is given the guideline of how to judge whether an answer is correct or not. The correctness of an answer is defined by two indicators: answerability and completeness. The answerability means the output from the system can answer the input question. The completeness means the content of the outputted answer covers the meaning of the gold answer. The two annotators made their judgments by comparing generated answers and gold answers. In the second step, the final judgment (correct or incorrect) was done based on the discussion of the two annotators (95\% agreement). The final score is accuracy which is the total number of correct answers divided by all gold answers.

\section{Preliminary results and discussion}\label{sec:result}
% Show our results of experiments.
This section shows the preliminary results of knowledge enhancement for LLMs. It first reports performance comparison of LLMs in context-aware and context-agnostic settings. It then shows an analysis based on actual outputs.

%It next reports the utilization of public knowledge bases to enhance the knowledge of LLMs

\subsection{Performance comparison}
As argued, LLMs need knowledge enhancement for working well in domain-specific tasks. To validate the argument, this section reports the results of LLMs for insurance QA in two settings. The first setting directly uses the knowledge of LLMs (context-agnostic - GPT). The second setting uses knowledge enhancement (context-aware). Note that the same indicators and questions were used for GPT for fair comparison (Algorithm \ref{lst:prompt_simple}, \ref{lst:prompt_rev_rulebook}, and \ref{lst:prompt_rev_kg}). This section also shows the contribution of table processing to improve the capability of LLMs for handling structured data.

\paragraph{Context-aware vs. context-agnostic}
%This section reports the comparison of the model using domain-specific knowledge enhancement (context) and the model without using the knowledge .
This section reports the results of GPT in two settings: context-aware and context-agnostic. To do that, the comparison was done with three models: (i) \textbf{GPT} with its knowledge using the prompt in Algorithm \ref{lst:prompt_simple}; (ii) \textbf{GPT+RB}: GPT with context from relevant rulebook's sections using the prompt in Algorithm \ref{lst:prompt_rev_rulebook}; and (iii) \textbf{GPT+RB+DBpedia}: GPT with context from relevant rulebooks' sections and DBpedia using the prompt in Algorithm \ref{lst:prompt_rev_kg}. More precisely, the context-agnostic setting includes the \textbf{GPT} model and the contex-aware setting contains the \textbf{GPT+RB} and \textbf{GPT+RB+DBpedia} models.

\begin{figure}[!h]
  \centering
  \subfigure[Results on the expert-created dataset.]{\includegraphics[width=0.44\linewidth]{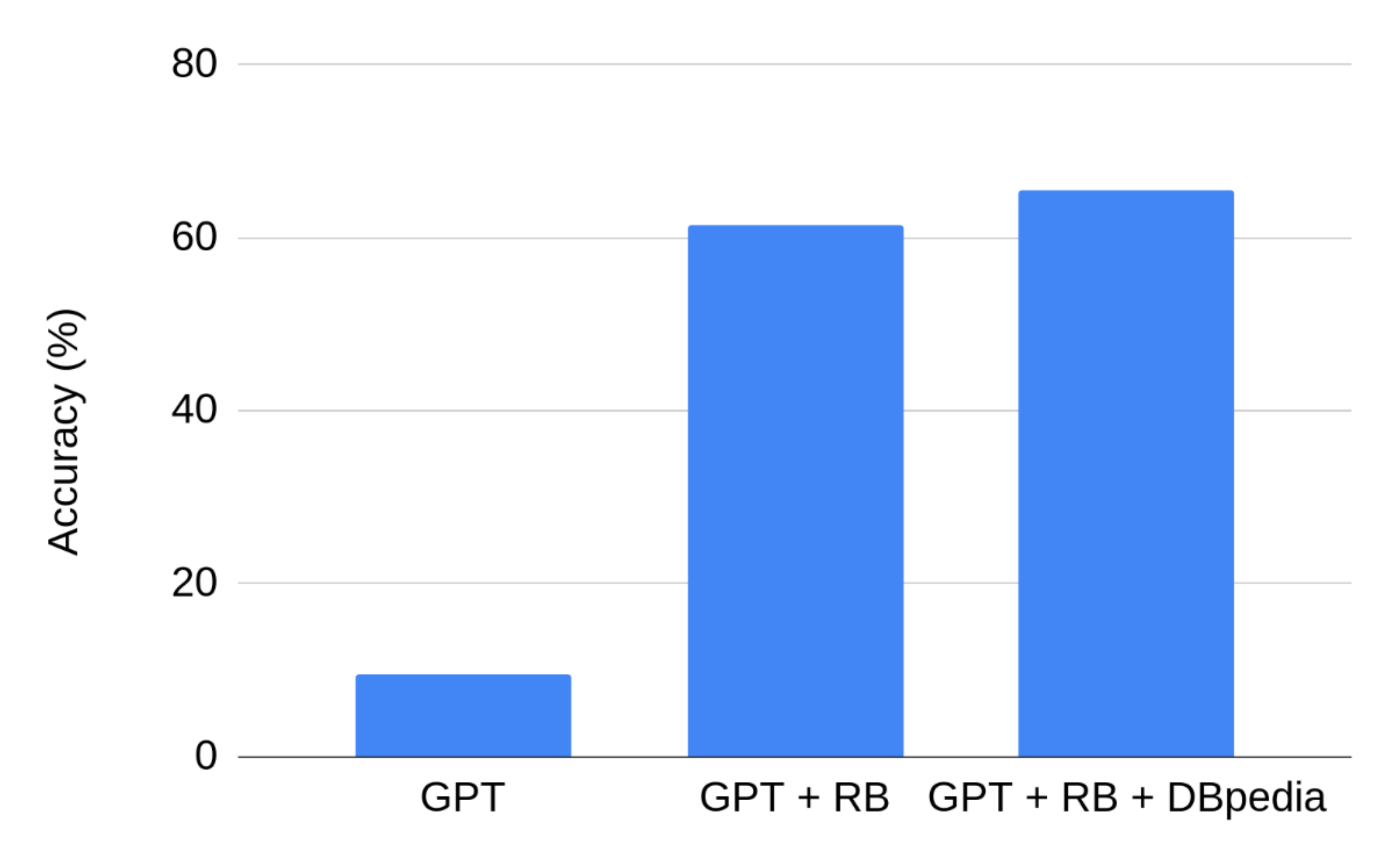}\label{fig:104}}
  \hspace{0.1\linewidth}
  \subfigure[Results on the synthesized dataset.]{\includegraphics[width=0.44\linewidth]{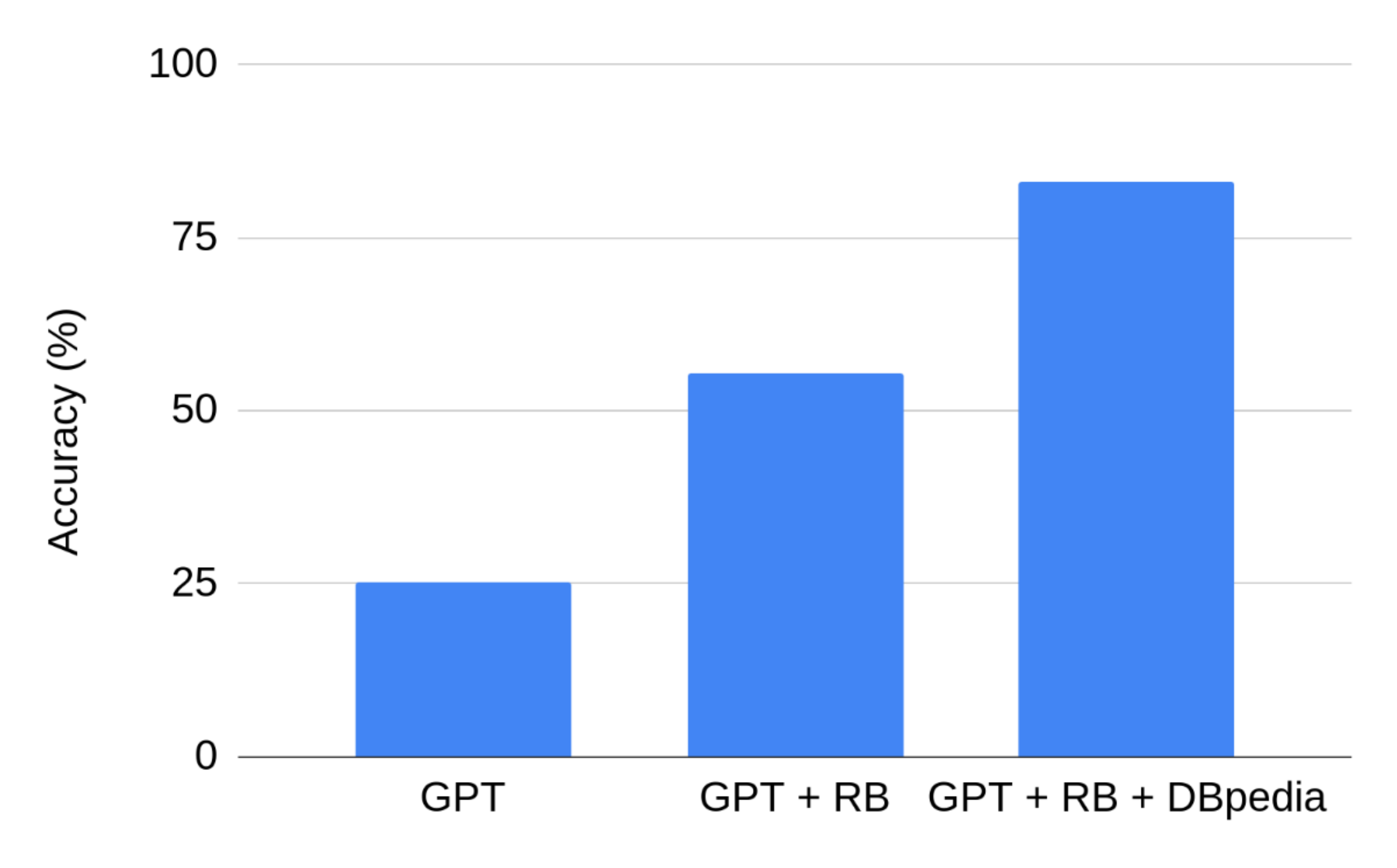}\label{fig:87}}
  \caption{Performance of GPT with different context using human evaluation.}
  \label{fig:context-no-context}
\end{figure}

Figure \ref{fig:context-no-context} shows the results of GPT across the two different settings.
The general trend indicates that knowledge enhancement is beneficial for GPT that can significantly improve the performance on the two datasets.
For the expert-created insurance dataset, there exists huge gaps between context-aware and context-agnostic settings. It is easy to understand that GPT's knowledge could not cover specific questions and concepts of insurance. It turns out that GPT can only answer 10 questions (9.60\%) over 104 questions. We observed that without knowledge enhancement, GPT tends to respond non-specific answers such as \textit{``Please provide more context..."} or \textit{``You should check their policies..."}. In contrast, domain-specific knowledge enhancement allows the GPT model to learn from a given context, such as a rulebook, and provide more correct answers. The GPT+RB+DBpedia model demonstrates a slight improvement compared to the GPT+RB model (65.40\% compared to 61.42\%). This improvement can be attributed to the utilization of DBpedia as a bridge between general knowledge and domain knowledge from the rulebooks. When relying solely on the rulebooks, GPT may struggle to understand certain relationships and connections between a question and its answer. For instance, without additional knowledge, GPT might not inherently recognize that \textit{diabetes} is considered as a lifestyle disease and should be covered under the lifestyle insurance policy. By incorporating DBpedia, GPT's understanding is enriched with supplementary information. DBpedia provides detailed explanation such as \textit{``Lifestyle diseases can be defined as diseases linked with one's lifestyle... which can lead to heart disease, stroke, obesity, type II diabetes"}. This additional knowledge allows GPT to make more informed connections and draw appropriate conclusions.

The results of the synthesized dataset share the same trend with the expert-created dataset, in which the performance of the context-aware setting is much better than that of the context-agnostic setting.
The direct use of GPT can only obtain 25.30\% accuracy. On the other hand, adding rulebooks and knowledge from DBpedia, GPT can significantly improve the performance of answering questions (55.42\% for the GPT+RB model and 83.13\% for the GPT+RB+DBpedia model).
It confirms the contribution of knowledge enhancement and validates our hypothesis in Section \ref{sec:method}.
There are two interesting points when observing Figure \ref{fig:104} and \ref{fig:87}. First, external knowledge, i.e., DBpedia plays a more important role in the synthesized dataset than the expert-created dataset. This is because, the synthesized dataset contains more questions that the rulebooks could not cover. So external knowledge helps to add additional information for the rulebooks. In contrast, almost questions in the expert-created dataset can be answered by only using the rulebooks. In this case, using external knowledge seems to be unnecessary.
In practice, domain-specific and common knowledge should be combined to provide comprehensive context for LLMs.
Second, the performance of GPT (context-agnostic) on the synthesized dataset is much better than that on the expert-created dataset. This may come from the creation process in which GPT can operate quite well on its generated questions.
The results in Figure \ref{fig:context-no-context} confirm the hypothesis stated in Section \ref{sec:method} that LLMs should not stand alone and it needs knowledge enhancement for working in domain-specific scenarios. Note that the results in Figure \ref{fig:context-no-context} already include table processing mentioned in the next section.

\paragraph{Handling tables}\label{sec:table_qa}
The rulebooks contain tables that make a challenge for GPT because GPT may not handle structured data well. So the tables need to be pre-processed to retain the columns and rows structure. After extracting, tables were represented in the form of sequences by concatenating text in each row and column. This representation helps GPT to understand the structure of tables and the relationships between individual cells within a table. By leveraging this approach, GPT can effectively process and analyze tabular data. Detailed information is shown in Algorithm \ref{lst:table_rep}.

\begin{figure}[!h]
    \centering
    \includegraphics[width=0.5\textwidth]{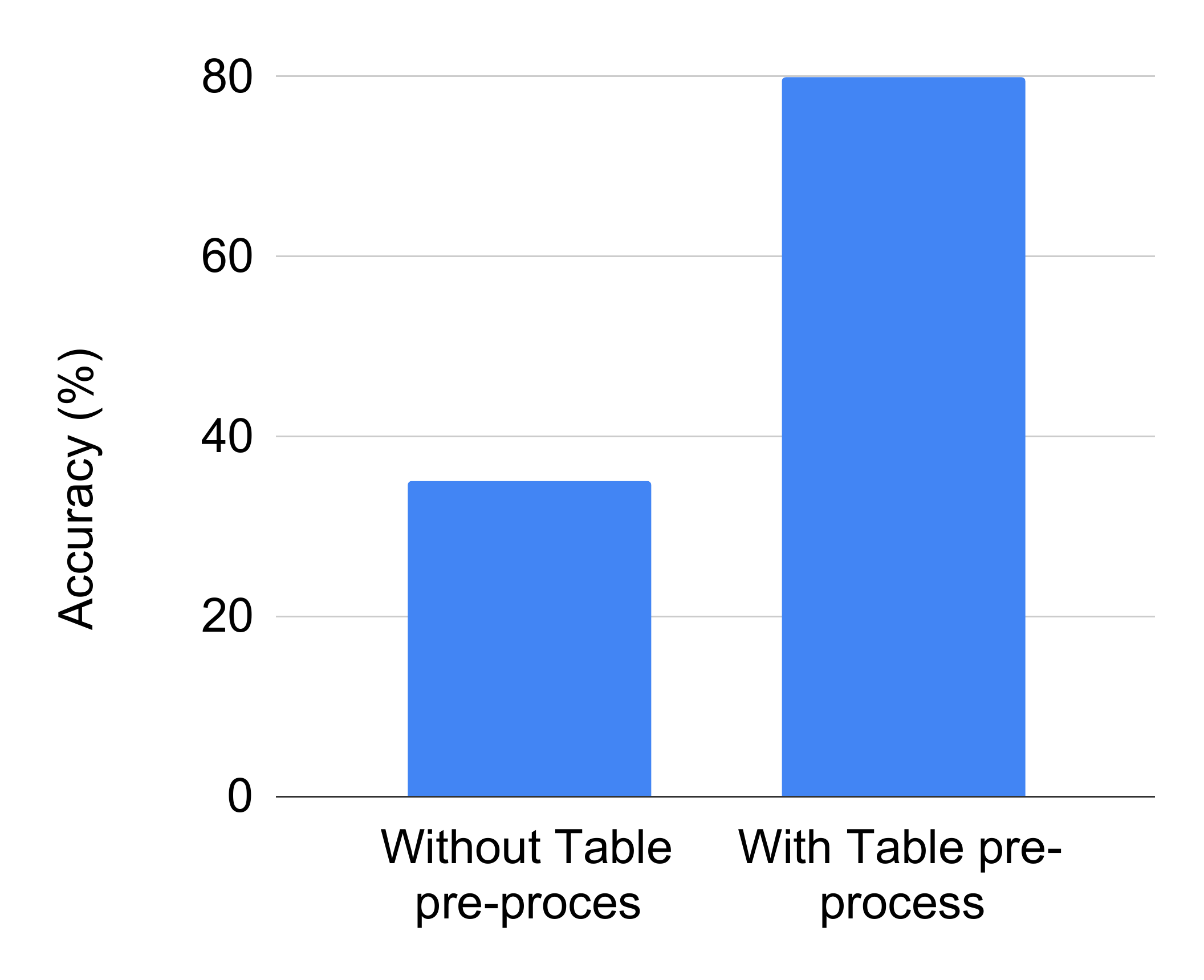}
    \caption{Performance of GPT with and without table processing on the expert-created dataset.}
    \label{fig:context_table}
\end{figure}

Figure \ref{fig:context_table} reports the comparison of GPT by using table processing. It indicates that employing table-like data as context improves the performance of GPT. This is because information stated in tables is required to correctly answer input questions. Given 20 questions related to data in tables, the model GPT+RB using table processing can answer 16 correct questions (80\%) while the model without using the processing can only answer 7 correct questions. It highlights the efficacy of structured data processing methods for the knowledge enhancement of LLMs. Note that we did not generate questions that require answer information in tables for the synthesized dataset.

%resulting in a substantially higher accuracy rate of 80\% compared to the performance of GPT without table-preprocessing. Specifically, the utilization of table data as context significantly improves the accuracy of GPT-based question-answering systems, highlighting the efficacy of this technique in enhancing the performance of language models in natural language processing tasks involving structured data.

\subsection{Questions need external knowledge for answering}\label{sec:external-help}

We observed that using external knowledge can help to answer input questions. This section shows the observation how external knowledge helps to improve the reasoning capability of GPT. To do that, we counted the number of questions that require additional knowledge for answering.

% \begin{table}[!h]
%     \centering
%     \caption{Performance of GPT+DBPedia and GPT+CoT. Heart disease, Stroke, Obesity, Diabetes, and Arthritis are lifestyle diseases. Avian flu, Ebola, Flu, Zika, and Rabies are non-lifestyle diseases, according to DBPedia (https://dbpedia.org/page/Lifestyle\_disease).}
%     \label{tab:i_prompt_kg_cot}
%     \begin{tabular}{|l|c|c|c|c|c|c|}
%         \toprule
%         \multirow{2}{*}{Disease} & \multicolumn{3}{c|}{GPT link diseases to lifestyle diseases} & \multicolumn{3}{c|}{Correct answer} \\ \cmidrule{2-7}
%         & GPT & GPT+RB & GPT+RB+DBPedia & GPT & GPT+RB & GPT+RB+DBpedia \\
%         \midrule
%          heart disease & no & yes & yes & no & yes & yes \\
%          stroke & no & yes & yes & no & yes & yes\\
%          obesity & no & yes & yes & no & yes & yes\\
%          diabete & no & yes & yes & no & yes & yes\\
%          arthritis & no & yes & yes & no & yes & yes\\
%         \midrule
%          avian flu & no & no & no & yes & yes & yes \\
%          ebola & no & no & no & yes & yes & yes\\
%          flu & no & no & no & yes & yes & yes\\
%          zika & no & no & yes & yes & yes & no\\
%          rabies & no & no & no & yes & yes & yes\\
%          \bottomrule
%     \end{tabular}
% \end{table}

\begin{figure}[!h]
  \centering
  \subfigure[Two questions need external knowledge for answering in the expert-created dataset.]{\includegraphics[width=0.44\linewidth]{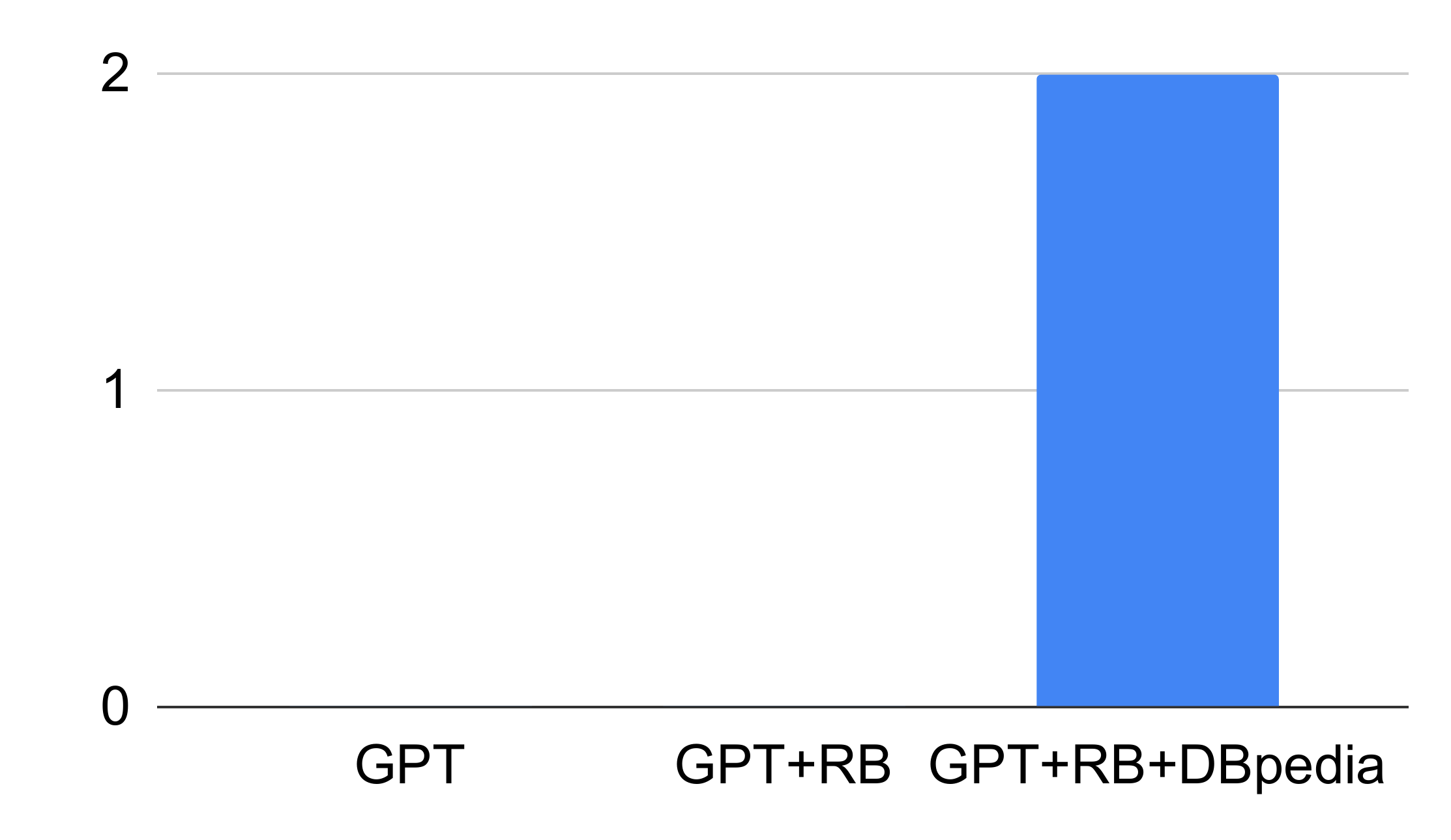}\label{fig:external_dai}}
  \hspace{0.1\linewidth}
  \subfigure[29 questions need external knowledge for answering in the sythesized dataset.]{\includegraphics[width=0.44\linewidth]{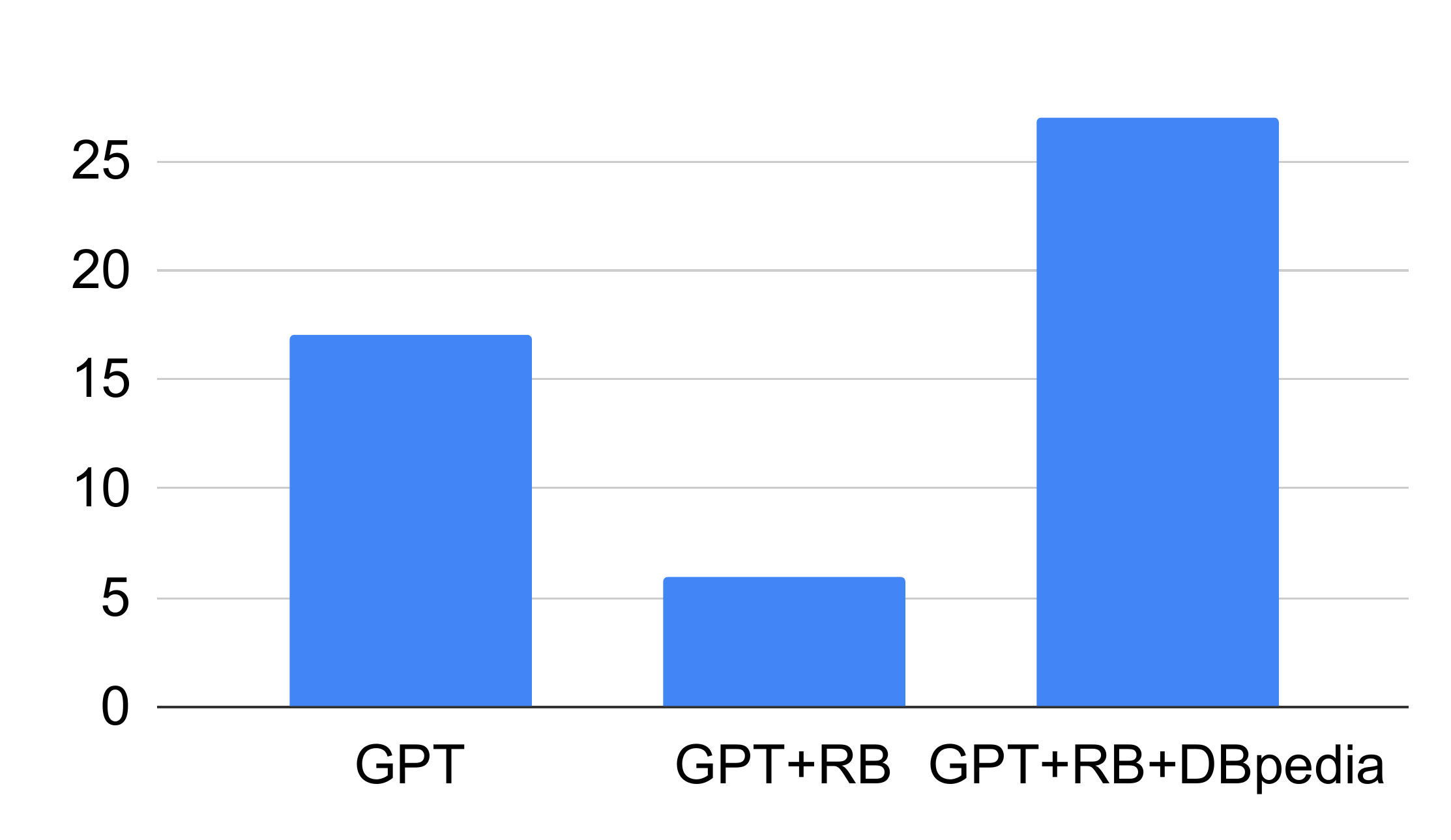}\label{fig:external_syn}}
  \caption{Performance of GPT on questions that require external knowledge.}
  \label{fig:external_performance}
\end{figure}

Figure \ref{fig:external_performance} shows that the number of questions that require external knowledge for answering in the expert-created dataset is much smaller than that in the synthesized dataset. For the first dataset, there only two questions that need external knowledge from rulebooks and DBpedia for answering. The GPT+RB+DBpedia model can answer these questions because retrieved information from DBpedia can cover entities that are not mentioned in the rulebooks. The GPT+RB and GPT could not answer the questions. Answering the two questions contributes 1.92\% of the improvement of GPT+RB+DBpedia in Figure \ref{fig:104}. In contrast, the synthesized dataset includes 29 questions that require external knowledge for answering. The GPT+RB+DBpedia can answer 27 questions while GPT can answer 17 questions. It confirms the significant improvement of the GPT+RB+DBpedia model compared to GPT.
Interestingly, the GPT+RB model can only answer six questions. A possible reason is that only using the rulebooks may not enough and may provide noisy context for input questions. However, the GPT+RB model can answer well other questions in Figure \ref{fig:87}.

%During experiments, there are cases in that GPT  fails to provide relevant response to a question that mentions a disease not present in the given context as shown in algorithm \ref{lst:hid_test}. Despite the LLM's capability to learn from a vast amount of knowledge, it still exhibits limitations in performing complex reasoning tasks. Based on this observation, two hypotheses have been proposed to explain such limitations. The first hypothesis suggests that GPT requires external knowledge sources, such as Knowledge Graphs, to supplement its understanding of a given context. This approach can provide additional contextual information to help a large language model comprehend the relationships between entities and concepts.

%Table \ref{tab:i_prompt_kg_cot} reports the results of the two hypotheses. It is clear that GPT cannot infer open-domain questions (lifestyle diseases that are not in the rulebooks) correctly. Meanwhile, complementing GPT with external DBPedia or extending its knowledge with CoT can significantly improve its ability to answer open-domain questions. There is a case where CoT considers the Zika disease as a lifestyle disease, but Zika must be a non-lifestyle disease. This might be the hallucination issue of LLMs. Overall, these experiments highlight the limitations of GPT and the potential benefits of supplementing its knowledge with external information sources.
%The proposed hypotheses provide valuable insights into the challenges of designing large language models and offer potential solutions to overcome their limitations.

\subsection{Output observation}\label{sec:output-observation}
Table \ref{tab:i_gpt_output_sample} shows answers of GPT under the three distinct models. GPT can answer general questions using its knowledge like ``\textit{What kind of surgery does a hysterectomy involve}?", ``\textit{What types of surgery are involved in breast reconstruction surgery}?", or ``\textit{If life or death of the insured person is unknown in whole life insurance, will the death benefit be paid}?" (Table \ref{tab:samples}). These questions do not require specific knowledge from policy rulebooks to answer. However, GPT fails to provide accurate responses when confronted with questions requiring specific information from policy rulebooks.
For example, for the first and second questions, GPT could not answer the question and requires to provide more context.
On the other hand, supplemented with additional knowledge from rulebooks and DBpedia, GPT is able to answer both general and specific questions with detailed explanations.
For the third and fourth questions, GPT can answer with common information. GPT+RB could not answer and require more context. GPT+RB+DBpedia can correctly answer the questions. This is because the questions contains entities that are not mentioned in the rulebooks. Therefore, retrieved information from DBpedia helps to enrich the information of the rulebooks.
GPT generates long outputs that do not directly answer the questions (row 1, 2, and 4) because the questions require additional knowledge and GPT tries to response as the nature of a generative LLM. In contrast, GPT with knowledge enhancement can answer correctly with detailed references.
\begin{table}[t]
\centering
\caption{\label{tab:i_gpt_output_sample}The outputs of GPT models by using different additional context. \textbf{(ii)} is only using GPT; \textbf{(ii)} stands for GPT+RB; and \textbf{(iii)} denotes GPT+RB+DBpedia.} 
\begin{tabular}{| p{0.2\linewidth} | p{0.1\linewidth} | p{0.6\linewidth} |}
  \toprule
  Question & Setting & Answer  \\
  \midrule
  \multirow{4}{1.0\linewidth}{How much is the radiation treatment benefit payment?} 
  & \textbf{(i)} & I'm sorry, as an AI language model, I don't have access to current information about radiation treatment benefit payment. Please provide more context or specify the location and time frame for a more accurate answer. \\
  & \textbf{(ii)} & The radiation treatment benefit payment is (Daily hospitalization amount) x 10 \\
  %& GPT+KG & The radiation treatment benefit payment is 10 times the daily amount of the hospitalization benefit \\
  & \textbf{(iii)} & The radiation treatment benefit payment is (Daily hospitalization amount) x 10 \\
  \midrule
  
  \multirow{4}{1.0\linewidth}{What is the maximum amount of advanced medical care benefits?} 
  & \textbf{(i)} & As an AI language model, I do not have access to specific information about a particular policy or insurance plan. Please provide more context or details about the policy or plan in question. \\
  & \textbf{(ii)} & The maximum amount of advanced medical care benefits is 20 million yen. \\
  %& GPT+KG & The maximum amount of advanced medical care benefits is 10,000 yen / The maximum amount of advanced medical care benefits is 20,000 yen.' \\
  & \textbf{(iii)} & The maximum amount of advanced medical care benefits is 20 million yen. \\
  \midrule
  
  \multirow{4}{1.0\linewidth}{Can I claim a second bone marrow donor benefit?} 
  & \textbf{(i)} & It is possible to claim a second bone marrow donor benefit, but it would depend on your specific insurance contract and the terms and conditions of your policy. \\
  & \textbf{(ii)} & The context provided does not contain information about a bone marrow donor benefit, so it is not possible to answer this question. \\
  %& GPT+KG & No, you cannot claim a second bone marrow donor benefit. The insurance policyholder is only eligible to receive the bone marrow donor benefit once. \\
  & \textbf{(iii)} & No, payment of the bone marrow donor benefit shall be made only once during the insurance period. (Article 10, Supplementary Regulations Concerning Payment of Bone Marrow Donor Benefits) \\
  \midrule
  
  \multirow{4}{1.0\linewidth}{What types of surgery are involved in breast reconstruction surgery?} 
  & \textbf{(i)} & Some common techniques used in breast reconstruction surgeries include breast implants, tissue expanders, flap reconstruction (taking tissue from another part of the body and using it to rebuild the breast), or a combination of these methods. \\
  & \textbf{(ii)} & The context does not provide a clear answer to this question. \\
  %& GPT+KG & Breast reconstruction surgery typically involves a variety of procedures, including tissue expansion, flap reconstruction, and implant reconstruction. \\
  & \textbf{(iii)} & Breast reconstruction surgery involves breast reconstructive surgery (Table 35) at a hospital or clinic (Table 6) during the insurance period. \\
  \bottomrule
\end{tabular}\vspace{-0.4cm}
\end{table}

\begin{figure}[!h]
  \centering
  \subfigure[Failure cases of the expert-created dataset.]{\includegraphics[width=0.43\linewidth]{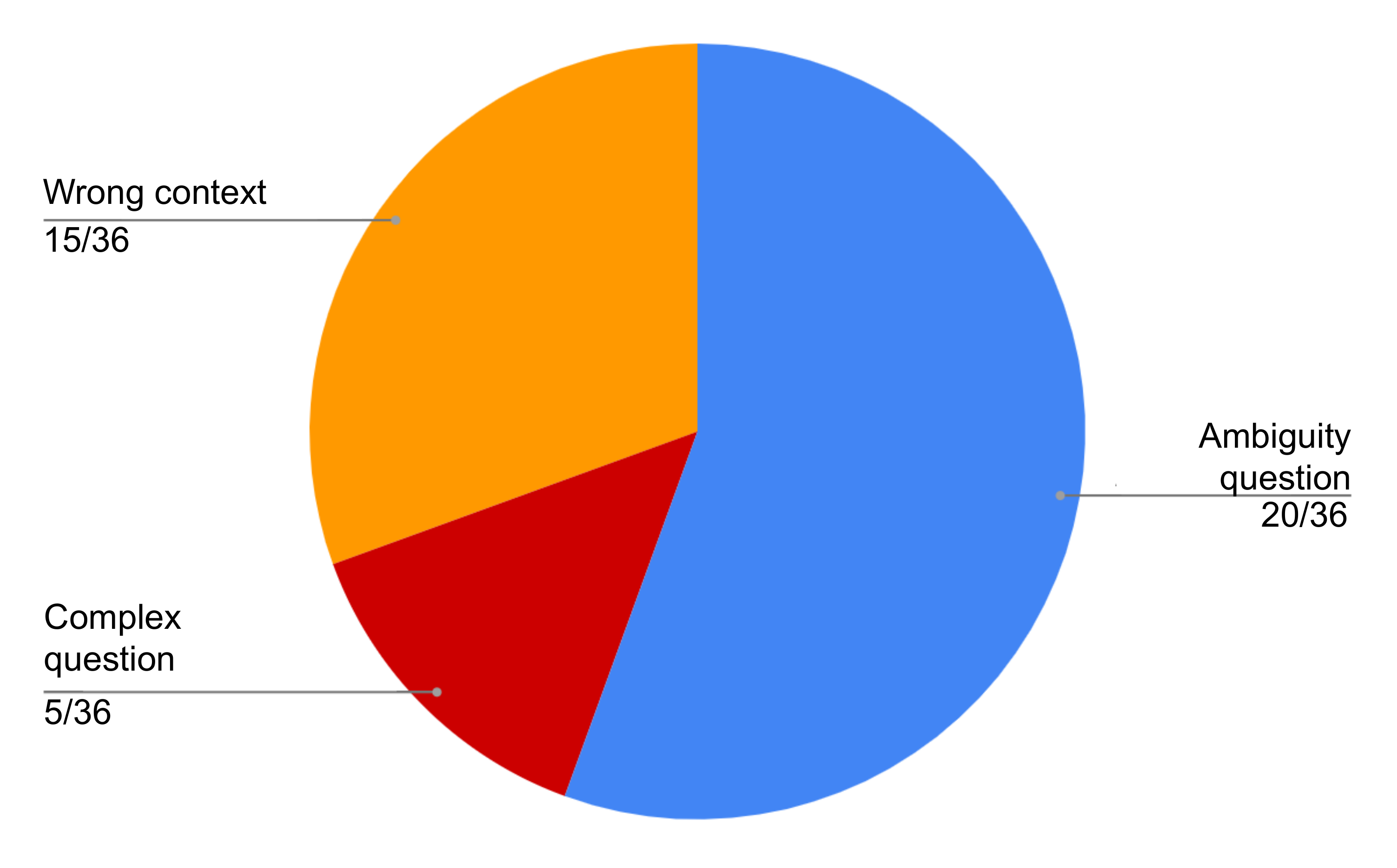}\label{fig:error_dis_ins}}
  \hspace{0.1\linewidth}
  \subfigure[Failure cases of the synthesize dataset.]{\includegraphics[width=0.43\linewidth]{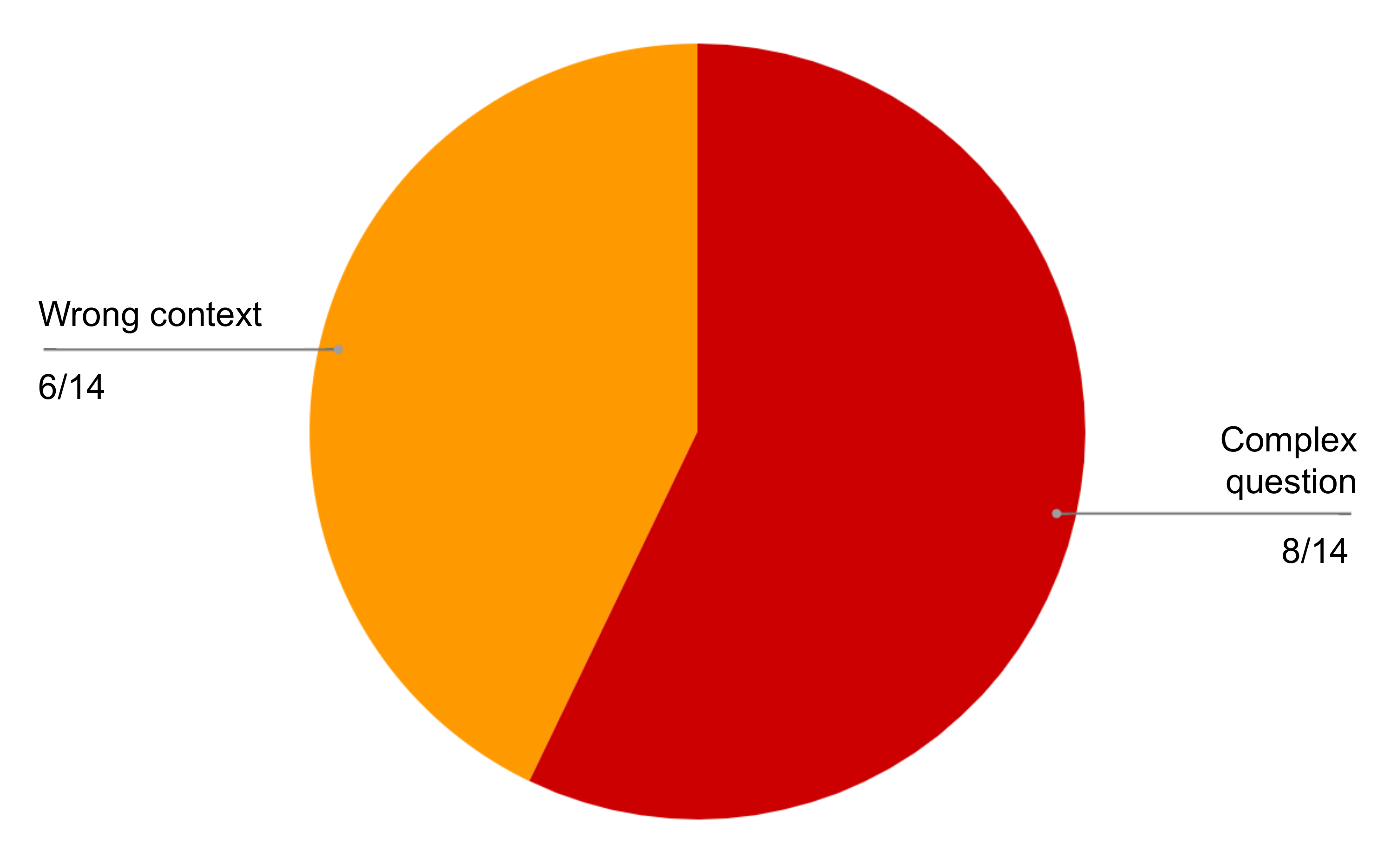}\label{fig:error_dis_syn}}
  \caption{Error distribution of GPT+RB+DBpedia of failure cases on the two datasets.}
  \label{fig:error_dis}
\end{figure}

Figure \ref{fig:error_dis} shows the failure analysis where GPT fails to provide accurate answers even after learning knowledge from relevant rulebooks and DBpedia (the best setting) on the two datasets. For the expert-created dataset, except for ambiguity cases, which take 55.55\% of wrong responses, 13.88\% of error cases are complex questions that GPT cannot infer directly in one-step reasoning. It may come from the nature of answering insurance questions that require multi-steps reasoning with deep understanding from domain experts. The remaining 41.66\% of error cases are caused by incorrect context retrieval. Our retrieval module uses a simple Cosine similarity method to identify the relevant context for a user's question. For the synthesized dataset,  57.15\% of error cases come from complex questions. This ratio is quite similar to the expert-created dataset. The ratio of wrong context retrieval is also large with 42.85\%. This shortcoming shares the same cause with the expert-created dataset in which the IR part uses a simple Cosine similarity method for relevant context retrieval. It suggests that there is a need to improve IR to reduce the incidence of errors in context retrieval by considering more sophisticated similarity and vectorization methods. Figure \ref{fig:error_dis} also shows that the expert-created dataset is more challenging than the synthesized dataset that has no ambiguity questions. A possible reason comes from the creation process of the synthesized dataset in which questions are usually generated by using one-step reasoning from GPT.

\section{Conclusion and Research Directions}\label{sec:conclusion}
% Show conclusion
This paper introduces an empirical analysis of adapting LLMs (GPT) to the QA task of the insurance domain. To do that, we design a model in which LLMs are empowered by using domain-specific knowledge. The knowledge enhancement provides an appropriate way that existing knowledge bases can be integrated into LLMs by using prompt-based learning methods. Preliminary results on two datasets in the insurance domain show four important points.
First, to create powerful AI applications, LLMs should not operate in isolation. Instead, their knowledge should be augmented for domain adaptation by integrating AI expertise with external domain-specific knowledge.
%LLMs should not stand alone to become powerful AI applications and LLMs' knowledge should be enhanced for domain adaptation.
Second, the combination of domain-specific knowledge, e.g., policy rulebooks and common knowledge, e.g., DBpedia benefits the inference of LLMs.
Third, table processing and understanding demonstrate the importance of table comprehension for achieving the ultimate end value of the target application.
%The observation also shows that public knowledge bases, e.g., DBPedia are beneficial for knowledge enhancement.
Finally, IR plays an important role that helps to retrieve relevant context for the inference of LLMs.
Our proposal provides an effective and a practical way that can adapt LLMs to business scenarios.

These insights suggest opening several research directions to prepare LLMs for industrial and real-world applications. One key area of focus could be using knowledge graphs containing open and closed-domain knowledge to enhance LLMs. By incorporating structured knowledge graphs into the model's training and inference processes, LLMs could better capture the relationships between entities and concepts, leading to more accurate and robust predictions. Another promising research direction is using chain-of-thought prompts to improve reasoning procedures. This approach involves providing a series of interconnected prompts to iteratively guide the model's reasoning process, allowing it to make more nuanced and sophisticated inferences. The possible final direction is to improve the IR part that helps to retrieve more relevant context for the reasoning of GPT.

% Show directions of utilizing LLMs

% \subsubsection*{Acknowledgments}
% We would like to thank Hajime Hotta for the idea of sharing the progress work of domain adaptation for LLMs and Jeff Yang for the discussion and comments on the draft version of the paper.

\bibliography{prompt-llm}
\bibliographystyle{iclr2023_conference}

\appendix
\section{Appendix}
\subsection{Sample prompts}

\begin{lstlisting}[floatplacement=hbt!,breaklines,
  frame=single,caption= A sample prompt without knowledge enhancement.,label=lst:prompt_simple]
prompt = ``Answer the question in a short and concise way: `When will I be notified of payment reason change?'"
\end{lstlisting}

\begin{lstlisting}[floatplacement=hbt!,breaklines,
  rulecolor=\color{black},
  frame=single,caption= A sample prompt using relevant rulebooks.,label=lst:prompt_rev_rulebook]
prompt = ``Answer the question in a short and concise way: `When will I be notified of payment reason change?', base on the context: `2. If the Company changes the provisions concerning the grounds for payment of advanced medical care benefits, etc. 
pursuant to the provisions of Paragraph 1, the Company shall notify the policyholder of such change at least two months prior to the date of such change.' "
\end{lstlisting}

\begin{lstlisting}[floatplacement=H,breaklines,
  rulecolor=\color{black},
  frame=single,caption=A sample prompt using the relevant information from the DBpedia KG.,label=lst:prompt_rev_kg]
prompt = ``Answer the question in a short and concise way based on the context and external information: ``He was hospitalized for a week due to diabetes. how much is his benefit amount her allowance?' ---Context: ``Hospitalization benefits for lifestyle-related diseases, Payment amount, The following amount per hospitalization (daily amount of hospitalization benefits) x number of days of hospitalization during the insurance period for treatment of lifestyle-related diseases listed on the left' ---External information: ``lifestyle disease | abstract | ... lead to heart disease, stroke, obesity, type II diabetes and lung cancer.... "    
\end{lstlisting}

\subsection{Prompts of using table data}

\begin{lstlisting}[floatplacement=H,breaklines,
  rulecolor=\color{black},
  frame=single,caption=Table representation for GPT.,label=lst:table_rep]
"Table name": [
  {
    ``row": name of row corresponding to the cell
    ``column": name of column coressponding to the cell
    ``value": content of cell
  },
]
Example:
``Women's Specific Insurance": [
    {
        ``row": ``Female Specific Surgery Benefits",
        ``column": ``Details of benefits",
        ``value": ``Surgery involving the breast, uterus"
    },
    {
        ``row": ``Breast Reconstruction Benefits",
        ``column": ``Details of benefits",
        ``value": "Breast reconstruction surgery for the breast"
    }
]    
\end{lstlisting}

% \subsection{Prompts of hidden knowledge, KG, CoT}

% \begin{lstlisting}[caption=Prompt template of GPT hidden knowledge testing., label=lst:hid_test,floatplacement=H,breaklines,
%   rulecolor=\color{black},frame=single]
% prompt = f``Answer the question and provide evidence for the Main question based on the table.
% -Table: {context}
% -Main Question: {ques}."
% \end{lstlisting}

% \begin{lstlisting}[floatplacement=H,breaklines,
%   rulecolor=\color{black},
%   frame=single,caption=Prompt template to use KG as context.,label=lst:prompt_kg]
% prompt = f``Answer the question and provide evidence for the Main question based on the table. Use knowledge graph as additional knowledge
% -Table: {context}
% -Knowledge graph: {e} | is sub class of | lifestyle related disease
% -Main Question: {ques}."
% \end{lstlisting}

% \begin{lstlisting}[floatplacement=H,breaklines,
%   rulecolor=\color{black},
%   frame=single,caption=Prompt template of Chain-of-Thought reasoning.,label=lst:prompt_cot]
% prompt = f``Answer the question in a short and concise way based on the table and knowledge graph 
% -Table: {context}
% -Main Question: {ques}.
% Before answer the Main question, answer these question.
%     1. Is there any disease or treatment or benefit or something else mentioned in the question
%     2. If yes, does it relate to the table.
%     3. If yes, what is the relationship.
%     Then answer and provide evidence for the Main question"
% \end{lstlisting}

\end{document}